\documentclass{article}

\usepackage{listings}
\usepackage{arxiv}
\usepackage{adjustbox}
\usepackage[utf8]{inputenc} 
\usepackage[T1]{fontenc}    
\usepackage{hyperref}       
\usepackage{url}            
\usepackage{booktabs}       
\usepackage{amsfonts}       
\usepackage{nicefrac}       
\usepackage{siunitx}        
\usepackage{microtype}      
\usepackage{lipsum}		
\usepackage{graphicx}
\usepackage{natbib}
\usepackage{doi}
\usepackage{bm}
\usepackage{rotating}
\usepackage{tikz}
\usepackage{caption,color,float}
\usepackage{multirow}       
\usepackage{url}            
\usepackage{booktabs}       
\usepackage{amsfonts}       
\usepackage{nicefrac}       
\usepackage{microtype}      
\usepackage{lipsum}
\usepackage{fancyhdr}       
\usepackage{graphicx}       
\usepackage{comment}        
\usepackage{amsmath, amssymb, bm}
\usepackage{subcaption} 

\usepackage{amsthm}%
\usepackage{mathrsfs}%
\usepackage[title]{appendix}%
\usepackage{xcolor}%
\usepackage{textcomp}%
\usepackage{manyfoot}%
\usepackage{algorithm}%
\usepackage{algorithmicx}%
\usepackage{algpseudocode}%
\usepackage{listings}%
\usepackage{adjustbox}
\usepackage{geometry}
\usepackage{tcolorbox}
\usepackage{siunitx}

\usepackage{longtable}
\usepackage{tabularx}
\usepackage{booktabs}
\usepackage{tabularx}
\usepackage{siunitx}
\usepackage{makecell}
\usepackage[bb=px]{mathalfa} 

\usepackage[varvw]{newtxmath}       
\usepackage{amsmath}
\mathchardef\mhyphen="2D 
\usepackage{booktabs}
\usepackage{multirow}
\usepackage{comment}
\usepackage{diagbox}
\usepackage{soul}
\usepackage{fancyhdr}
\usepackage{wrapfig}
\usepackage{graphicx}

\usepackage{subcaption}

\title{Sensing Without Colocation: Operator-Based Virtual Instrumentation for Domains Beyond Physical Reach}

\author{
        {Jay Phil ~Yoo} \\
    Nuclear, Plasma \& Radiological Engineering \\
    University of Illinois at Urbana-Champaign \\
    Urbana, IL 61801, USA \\
    \texttt{jayyoo2@illinois.edu} \\
    \And
    {Kazuma ~Kobayashi} \\
    Nuclear, Plasma \& Radiological Engineering \\
    National Center for Supercomputing Applications \\
    University of Illinois at Urbana-Champaign \\
    Urbana, IL 61801, USA \\
    \texttt{kazumak2@illinois.edu} \\
    \And
    {Souvik ~Chakraborty} \\
    Department of Applied Mechanics\\
    Yardi School of Artificial Intelligence\\
    Indian Institute of Technology Delhi\\
    New Delhi, India \\
    \texttt{souvik@am.iitd.ac.in} \\
    \And
    {Syed Bahauddin ~Alam} \\
    Nuclear, Plasma \& Radiological Engineering \\
    National Center for Supercomputing Applications \\
    University of Illinois at Urbana-Champaign \\
    Urbana, IL 61801, USA \\
    \texttt{alams@illinois.edu} \\
}

\renewcommand{\headeright}{Preprint}
\renewcommand{\undertitle}{Preprint}
\renewcommand{\shorttitle}{\textit{Sensing Without Colocation: Operator-Based Virtual Instrumentation
}}

\hypersetup{
pdftitle={A template for the arxiv style},
pdfsubject={q-bio.NC, q-bio.QM},
pdfauthor={David S.~Hippocampus, Elias D.~Striatum},
pdfkeywords={First keyword, Second keyword, More},
}

\begin{document}
\maketitle

\begin{abstract}

Classical sensing rests on one foundational assumption: the quantity of interest must be colocated with the measurement device. This is not an engineering convenience. It is the organizing principle of every instrumentation standard developed over the past century, and it fails completely at aviation altitude, where no physical sensor can survive long enough to monitor the cosmic radiation field that irradiates millions of aircrew annually. We establish that this barrier is resolved by a new sensing principle: when the sensor manifold and the target field manifold are physically disjoint, a learned operator bridging them \emph{is} the instrument. We term this \textbf{operator-theoretic virtual sensing} and instantiate it in \textbf{STONe}, which maps \textbf{twelve} ground-based neutron monitors (sparse, indirect, surface-bound) to the complete global dose field at 10{,}000\,m across \textbf{180-day} horizons, achieving sub-millisecond inference where Monte Carlo transport requires hours. Deployed without modification on an NVIDIA Jetson Orin Nano embedded AI platform at 7.3\,W average system power and 143.3\,MB GPU memory footprint; within the envelope of photovoltaic-powered field hardware co-locatable with existing neutron monitor stations, STONe constitutes a physically realizable sensing device of a new category: an instrument whose measurement principle is operator-theoretic and whose deployment constraint is the power budget of remote environmental monitoring infrastructure, not the accessibility of the target domain.

\end{abstract}

%
{ 
\section{Introduction}
Galactic Cosmic Radiation (GCR) presents a sensing problem no hardware 
advance can resolve. At commercial flight altitudes (10--12\,km), dose 
rates reach 50--100 times sea-level values and no operational sensor 
exists at altitude~\cite{sigurdson2004cosmic, silverman2009medical}: 
the physical environment precludes permanent deployment at global scale. 
Twelve ground-based neutron monitors provide the only available 
observations, while the dose field at 10{,}000\,m governing exposure 
remains beyond the reach of any transducer.

Conventional sensing assumes that the quantity of 
interest is colocated with, or continuously connected to, the measurement 
device. That assumption fails here by construction: ground instruments measure 
secondary neutron flux at the surface, while the public health quantity of 
interest is the effective dose field at 10{,}000\,m, a physically distinct and 
inaccessible regime. The coupling between these two domains is nonlinear, 
nonlocal, and mediated by primary cosmic ray spectra, geomagnetic modulation, 
and atmospheric transport processes that are never directly observed. 
High-fidelity Monte Carlo simulations can predict the dose field but require 
hours to days per evaluation~\cite{palmer2019ecmwf}, far too slow for 
operational deployment.
This constitutes a fundamental instrumentation failure: while surface-based sensor networks exist, no sensing principle converts their output into real-time, altitude-resolved atmospheric data across global airspace.

Traditional data-driven approaches (POD, 
DMD)~\cite{schmid2010dynamic, tu2013dynamic, brunton2022data} or equation 
discovery (SINDy, PDE-FIND)~\cite{brunton2016discovering, rudy2017data} 
cannot resolve the nonlinear, nonlocal coupling that governs GCR cascades, as they presume explicit governing equations in known coordinates. The assumption fails when the mapping from ground observables to high-altitude dose fields 
involves latent physical processes that remain unobserved and cannot be 
parameterized in closed form.

Neural operator frameworks represent a conceptually distinct step: learning 
solution operators that map input functions to output functions without 
requiring explicit differential operators. Deep Operator Networks 
(DeepONet)~\cite{lu2021learning} and Fourier Neural Operator 
(FNO)~\cite{li2020fourier} have demonstrated real-time predictive capability 
in nuclear and multiphysics systems~\cite{kobayashi2024improved, 
hossain2025virtual, kobayashi2024deep}. However, in existing applications 
these architectures function as fast surrogates for known simulators, not as 
a sensing principle in which the operator itself constitutes digital 
instrumentation that replaces unavailable measurements. This distinction is fundamental, reframing operator learning from a computational accelerator to a novel measurement modality.

For time-dependent systems, autoregressive formulations propagate states 
step-by-step~\cite{lu2021learning, li2020fourier}, accumulating error over 
long horizons~\cite{diab2025temporal, nayak2025ti}. More critically, 
autoregressive models inherit a hidden sensing assumption: that the future 
state inhabits the same coordinate space and physical domain as the input. 
That assumption is violated by construction in cross-domain virtual sensing, 
where the input manifold (ground surface) and the target manifold 
(10{,}000\,m altitude) are physically disjoint. Temporal physics-integrated 
models such as TI(L)-DeepONet~\cite{nayak2025ti} embed numerical integrators 
but remain domain-coincident. Sequence-to-sequence architectures eliminate 
iterative feedback~\cite{brandstetter2022message, gupta2022towards, 
lippe2023pde} but do not address the cross-manifold sensing problem. No 
existing framework provides a principled, stable operator for mapping sparse, 
indirect surface observations to inaccessible field states over operationally 
meaningful long-term horizons.

A parallel transformation in weather prediction illustrates both the promise 
and the gap. Models such as FourCastNet~\cite{pathak2022fourcastnet} and 
GraphCast~\cite{scarselli2008graph, lam2023learning} achieve global-scale 
predictive fidelity through learned operators rather than numerical 
integration~\cite{bi2023accurate, kochkov2024neural, price2025probabilistic}. 
Yet each system assumes input and output fields share the same spatial grid 
and physical domain. For radiation dose sensing, that assumption does not 
hold: the sensor manifold (twelve irregularly distributed ground stations) 
and the target manifold (global dose field at 10{,}000\,m) are non-coincident 
in both geometry and physics. This constitutes a fundamental structural gap that transcends preprocessing and reflects the absence of a sensing framework capable of stably bridging disjoint physical regimes at global scale over extended horizons.

Motivated by this gap and enabled by multi-decadal neutron monitor archives, 
mature sequence modeling architectures, and modern AI hardware, we propose a 
non-autoregressive neural operator framework designed specifically for virtual 
sensing across physically disjoint domains. The approach learns a direct 
sequence-to-sequence operator mapping sparse, irregularly distributed ground 
sensor histories to the complete future spatiotemporal dose field at 
10{,}000\,m in a single forward pass. Rather than forecasting, the objective 
establishes a new instrumentation principle in which learned operators 
bridging accessible and inaccessible manifolds constitute the measurement 
apparatus itself, representing a novel sensing modality.

The cross-domain operator is well-posed under the physical 
conditions present in this system. Ground neutron monitors 
provide long-term temporal observability of solar modulation 
cycles spanning years to decades, ensuring that the branch 
input encodes the slowly varying GCR forcing that drives 
altitude-dependent dose variation. The atmospheric transfer 
function relating surface neutron flux to aviation-altitude 
dose is physically stable over the sensing horizon: it is 
governed by geomagnetic cutoff rigidity and atmospheric 
depth, both of which vary smoothly and predictably. The 
operator is therefore identifiable from the available 
observations, and the inverse sensing problem is 
well-conditioned over the 180-day horizons considered here. 
This distinguishes operator-theoretic virtual sensing from 
generic inverse problems where identifiability cannot be 
assumed.

Classical sensing hardware couples a transduction element 
to a fixed signal-conditioning algorithm; the physical 
package and its embedded computation are inseparable---
together they constitute the instrument. We adopt this 
architectural definition and extend it: when the 
transduction element is replaced by a learned operator 
embedded in physical hardware, the resulting device is an 
instrument in the same structural sense. The novelty lies in the sensing principle: the operator-theoretic measurement of a 
physically inaccessible field from a deployable embedded 
device. This distinction is what separates the present 
work from prior operator learning applications, which 
function as computational accelerators rather than as 
sensing instruments that replace unavailable measurements.

\vspace{-1mm}
\begin{tcolorbox}[colback=blue!10!white, colframe=blue!50!black, 
coltitle=white, fonttitle=\bfseries]
The proposed \textbf{Spatio-Temporal Operator Network (STONe)} extends the 
Sequential Deep Operator Network (S-DeepONet)~\cite{he2024sequential} into 
a new regime of \textbf{operator-based digital instrumentation} for 
\textbf{cross-domain virtual sensing}. Rather than treating operator learning 
as surrogate modelling, STONe formalizes sensing as a learned operator that 
maps sparse, indirect measurements to inaccessible field states, thereby 
\textbf{replacing dense physical sensors} when deployment is impractical. 
STONe combines recurrent and attention-based encoders with a coordinate-conditioned trunk to infer complete spatiotemporal field trajectories from heterogeneous sensor streams. Its 
\textbf{non-autoregressive design} supports \textbf{long-horizon stability} 
by eliminating iterative error feedback, while its \textbf{operator 
decomposition} resolves domain mismatch natively, enabling stable inference 
across physically disjoint manifolds. In our setting, STONe transforms 
\textbf{severe sparsity} (12 ground stations) into \textbf{global} dose 
fields at \textbf{10{,}000\,m} over \textbf{180-day} horizons, establishing 
a \textbf{new sensing principle}: computational operators as virtual sensors 
for inaccessible environments with operational-scale coverage.
\end{tcolorbox}
\vspace{-1mm}

\vspace{-1mm}
\begin{tcolorbox}[colback=gray!10!white, colframe=blue!50!black, 
title=\textbf{Key contributions:}, coltitle=white, fonttitle=\bfseries]
\begin{itemize}

    \item \textbf{Learned operator sensing principle:}
    STONe formalizes sensing as a learned operator 
    $\mathcal{G}: \mathcal{H}_s \!\rightarrow\! \mathcal{H}_t$
    mapping sparse, indirect measurements to inaccessible fields across 
    disjoint manifolds $(\mathcal{H}_s \cap \mathcal{H}_t = \emptyset)$,
    establishing a fundamentally new instrumentation class.

    \item \textbf{Operational-scale virtual sensing under extreme constraints:}
    From \textbf{12} ground stations, STONe reconstructs \textbf{global} 
    dose fields at \textbf{10{,}000\,m} with \textbf{180-day} stability 
    under the measurement constraints of the deployed NMDB network.

    \item \textbf{Real-time deployability:}
    Sub-millisecond inference, orders-of-magnitude faster than 
    conventional transport solvers, positions STONe as deployable 
    digital instrumentation rather than a surrogate solver.

    \item \textbf{Physical instrument realization:}
    Deployed on a Jetson Orin Nano at 7.3\,W and 43.5\,ms per 
    complete 180-day global rollout, without retraining or 
    hardware-specific modification, demonstrating physical 
    realizability within the power envelope of field-deployable 
    instrumentation.

    \item \textbf{Universality beyond radiation:}
    The operator formulation generalizes to any domain where sparse, 
    indirect measurements must inform inaccessible fields, including 
    reactor monitoring, structural health diagnostics, and subsurface 
    systems.

\end{itemize}
\end{tcolorbox}
\vspace{-1mm}

\begin{figure}[!htbp]
    \centering
    \includegraphics[width=1.1\linewidth]{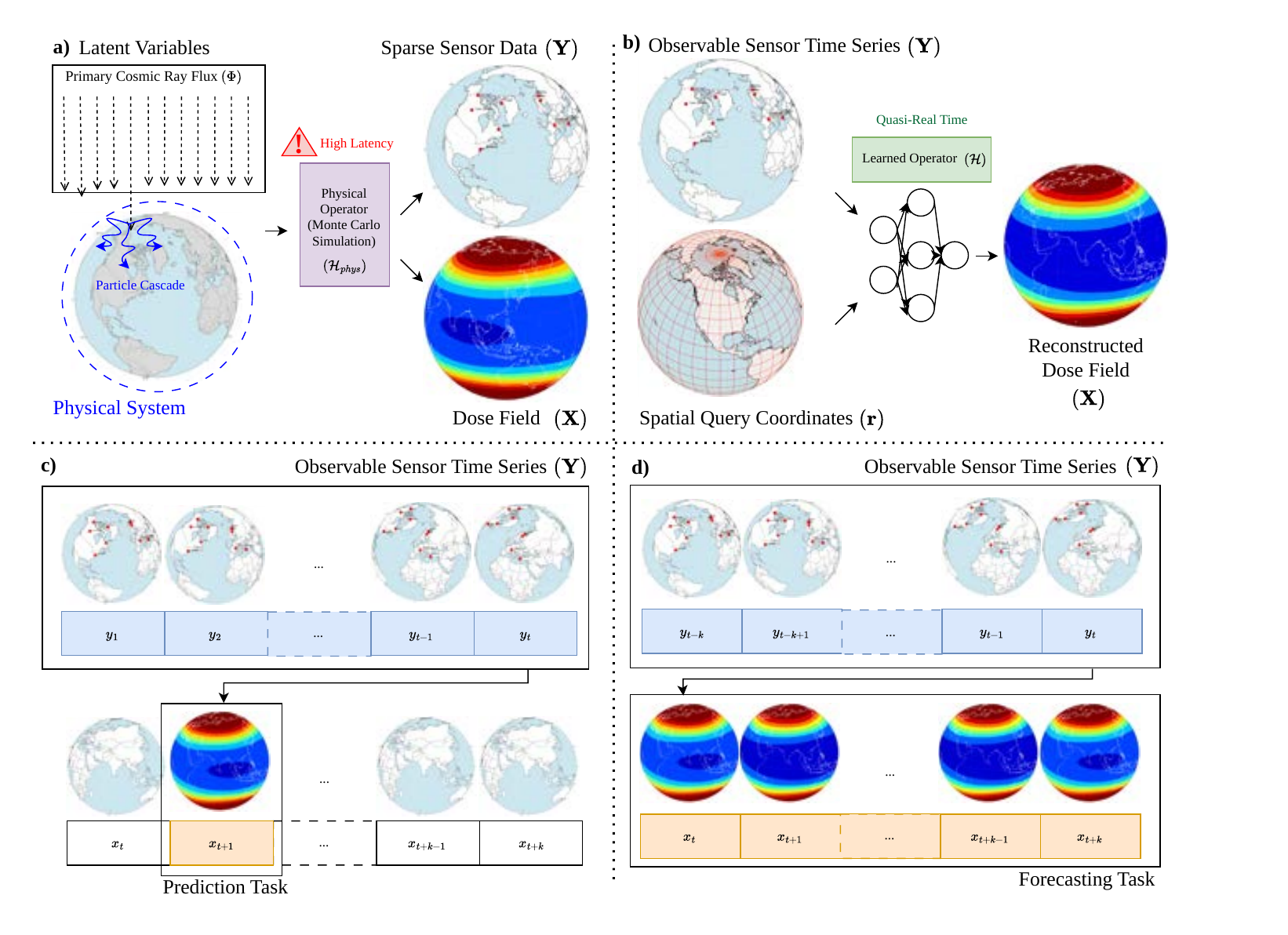}
    \caption{\textbf{Conceptual framework for virtual sensing: extending ground-based neutron monitor networks to aviation altitudes through learned operators.} The figure illustrates the transition from traditional physics-based forward modeling to a learned operator that creates virtual radiation sensors throughout the atmospheric column. \textbf{a)} Forward problem (physics-based simulation): Unobservable latent drivers, primarily the primary cosmic-ray flux ($\Phi$), interact with Earth's atmosphere and geomagnetic field. High-fidelity Monte Carlo transport models particle cascades, yielding sparse ground-based neutron monitor measurements ($\mathbf{Y}$) and high-altitude radiation dose fields ($\mathbf{X}$) where physical sensors cannot be deployed. Evaluating the latent physical operator $\mathcal{H}_{\text{phys}}$ incurs prohibitive computational latency for real-time applications. \textbf{b)} Learned virtual sensing operator: A neural operator $\mathcal{H}$ extends the sparse ground sensor network by mapping observable neutron monitor time series ($\mathbf{Y}$) and spatial query coordinates ($\mathbf{r}$) directly to high-resolution virtual dose measurements $\mathbf{X}$ at aviation altitudes, creating a dense virtual sensor network for operational monitoring. \textbf{c)} Single-step prediction (one-step ahead): Given a sequence of ground sensor inputs $\mathbf{Y}_{1:T}$ and spatial queries $\mathbf{r}$, a predictor outputs virtual dose measurements $\mathbf{X}_{t+1}$ at the next time step. \textbf{d)} Full-rollout virtual sensing (this work): A non-autoregressive operator predicts the entire future virtual sensor field sequence $\mathbf{X}_{1:K}$ from sparse ground measurements $(\mathbf{Y}, \mathbf{r})$ in a single pass (solid arrows), enabling stable, long-range atmospheric column monitoring.}
    \label{fig:problem}
\end{figure}
\vspace{-1mm}

\section{Methodology}
\subsection{Data Preparation and Preprocessing}
Accurately estimating cosmic radiation dose is critical for aviation safety and space weather monitoring, but traditional physics-based simulations introduce latency for real-time applications. To address this, we reframe the task as a data-driven inverse problem~\cite{kobayashi2025proxies}. Instead of simulating the complex chain of events from primary cosmic rays to radiation dose, our model learns a direct mapping from sparse, ground-based neutron sensor measurements to high-resolution global dose fields (Figure \ref{fig:problem}) solving the inverse problem.

The forward modeling approach simulates the causal chain from primary cosmic ray fluxes ($\Phi$) to downstream effects like ground-level neutron counts and dose rates. For a given sensor $n$, the neutron count $Y_n$ is a function of the primary flux, atmospheric conditions ($A$), the geomagnetic field ($G$), and altitude ($h$):
\begin{equation}
    Y_n = f_n(\Phi, A, G, h) \in \mathbb{R}^{T}
\end{equation}
where $f_n$ represents the complex particle transport physics and $T$ is the number of time steps. For a network of $S$ sensors, the combined data form a matrix $\mathbf{Y} = \{ Y_1, \dots, Y_N \} \in \mathbb{R}^{T \times N}$.

Similarly, the effective dose rate $X_j$ at a specific location $j$ is determined by the same physical drivers, modeled by a function $g_j$:
\begin{equation}
    X_{j} = g_{j}(\Phi, A, G, h)
\end{equation}
The collection of dose rates across $P$ evaluation points forms the dose field $\mathbf{X} = \{ X_1, \dots, X_P \} \in \mathbb{R}^{P}$. This defines a latent physical operator $\mathcal{H}_\text{phys}:(\Phi, A, G, h) \to \mathbf{X}$. While physically grounded, this forward approach is computationally expensive due to the need for extensive Monte Carlo transport simulations, making it infeasible for real-time global monitoring (Figure \ref{fig:problem}a).

To overcome the limitations of forward modeling, we formulate an inverse problem that directly maps observable proxies (i.e., neutron counts on the ground) to the desired dose fields. Instead of simulating latent variables like $\Phi$, the model learns a spatiotemporal operator $\mathcal{H}$ that reconstructs the dose field $\mathbf{X}$ from the observed neutron monitor time series $\mathbf{Y}$ and a set of spatial query coordinates $\mathbf{r}$. We define this learned inverse operator as:
\begin{equation}
    \mathbf{X} = \mathcal{H}(\mathbf{Y}, \mathbf{r})
\end{equation}
where $\mathbf{Y} \in \mathbb{R}^{T \times N}$ is the input from the neutron monitors and $\mathbf{r} \in \mathbb{R}^{P}$ represents the spatial query locations. This formulation bypasses the explicit simulation of intermediate physical processes, enabling fast, resolution-agnostic reconstruction of the global radiation dose distribution from real-world sensor data (Figure \ref{fig:problem}b). The neutron counting data used in this study was obtained from the Neutron Monitor Database (NMDB) \cite{mavromichalaki2011applications,nmdb}.

Notably, forecasting models that are designed for problems where the input and target fields share a common spatial domain cannot be directly applied to this problem. The problem setup involves a fundamental domain mismatch: sparse, irregularly distributed ground-based measurements as input, versus dense, global radiation dose fields at high altitudes as output. This domain mismatch, combined with the incomplete spatial coverage of the observational network, violates the assumptions of structured, grid-based inputs covering the entire target domain.

In addition, the problem is designed to solve for a sequence-to-sequence solution. In autoregressive schemes, a model iteratively consumes its own predictions as input, which leads to compounding errors over long horizons. In contrast, our non-autoregressive formulation predicts the entire future trajectory in a single pass, thereby mitigating error accumulation (Figure~\ref{fig:problem}d).
Namely, we learn an operator $\mathcal{H}$ that maps an input sequence of sensor observations $\mathbf{Y} \in \mathbb{R}^{T \times N}$ and a set of spatial query coordinates $\mathbf{r} \in \mathbb{R}^{P}$ to the full sequence of future dose fields $\mathbf{X}_{1:K} \in \mathbb{R}^{P \times K}$:
\[
    \mathbf{X}_{1:K} = \mathcal{H}(\mathbf{Y}, \mathbf{r}),
\]
where $K$ denotes the number of leading forecast steps.

The methodology employs an effective dose dataset~\cite{iwamoto2022benchmark} calculated using EXPACS~\cite{sato2015analytical, sato2016analytical, expacs} for an altitude of 10{,}000~m at a 1$^{\circ}$ latitude--longitude resolution, replicating environmental conditions relevant to aviation safety and space weather monitoring. The dataset is generated at a daily temporal resolution, providing dose estimates under varying geomagnetic and atmospheric conditions. Following the EXPACS code manual \cite{expacs}, the water fraction parameter was set to 0.15, representing the recommended setting. This dataset is designed to evaluate the model's ability to reconstruct high-dimensional spatiotemporal fields from sparse measurements, which are characteristic of the complex dynamics of cosmic radiation dose.

The dataset is structured into three primary components to facilitate operator learning. The \textit{Branch Input} consists of time-series measurements of neutron counts derived from a global network of 12 sparse, ground-based neutron monitors. These measurements, spanning the period from 2001 to 2023, serve as indirect proxies for the cosmic radiation field at each time step. Concurrently, the \textit{Trunk Input} provides the spatial query points for field reconstruction, defined by global grid coordinates normalized to the interval $[0,1]$ for both latitude and longitude, thereby offering a continuous representation of the spatial domain. The \textit{Target Output}, serves as the ground-truth reference for model evaluation and is composed of the full-field radiation dose rates across the global grid corresponding to each time step.

Data preprocessing involves the construction of input-output pairs using a sliding window approach to capture temporal dependencies. A temporal window of a fixed size, $T=K=180$, is applied to the neutron counts time series, generating sequences of input observations that are paired with the corresponding future radiation dose fields over an identical prediction horizon steps. The dataset is subsequently partitioned into training (45\%), validation (10\%), and testing (45\%) subsets. Crucially, the test set is reserved for evaluating the model's generalization to unseen temporal regimes. This chronological split preserves temporal continuity, ensuring that the model is assessed under realistic conditions where future states are not available during training. This preprocessing strategy is designed to facilitate robust training and evaluation of the STONe model for its designated tasks of long-term forecasting and spatiotemporal field reconstruction.

\subsection{DeepONet Framework}
The DeepONet is a neural network architecture designed to learn nonlinear operators, which are mappings between infinite-dimensional function spaces, $\mathcal{G}: \mathcal{I} \rightarrow \mathcal{O}$ \cite{lu2021learning}. The theoretical foundation for this architecture is the universal approximation theorem for operators, which states that a neural network can approximate any nonlinear continuous operator to arbitrary accuracy \cite{chen1995universal}.

A DeepONet approximates the evaluation of an output function $\mathcal{G}(\bm{u})$ at a specific query location $\bm{r} \in \Omega \subset \mathbb{R}^d$. The architecture achieves this through two distinct subnetworks. The first is the \textbf{Branch Network}, which encodes the input function $\bm{u} \in \mathcal{I}$ by evaluating it at a fixed set of $m$ sensor locations $\{\bm{x}_j\}_{j=1}^m \subset \Omega$. This network processes the discrete representation, $[\bm{u}(\bm{x}_1), \dots, \bm{u}(\bm{x}_m)]$, to produce a latent vector representation $b(\bm{u}) \in \mathbb{R}^q$. The second subnetwork is the \textbf{Trunk Network}, which encodes the continuous domain of the output function by taking a query coordinate $\bm{r} \in \Omega$ as input and mapping it to a corresponding basis vector $t(\bm{r}) \in \mathbb{R}^q$. The operator's output at the location $\bm{r}$ is then approximated by the inner product of the outputs from these two networks, often with an added bias term $\beta$:
\begin{equation}
    \mathcal{G}_\theta(\bm{u})(\bm{r}) = \sum_{i=1}^q b_i(\bm{u}) t_i(\bm{r}) + \beta \approx \mathcal{G}(\bm{u})(\bm{r}),
    \label{eq:deeponet}
\end{equation}
where $\theta$ denotes the set of all trainable parameters. This architecture learns separate representations for the input function and the output evaluation coordinates. This separation allows the model to make predictions at any location, independently from the training data's resolution and domain. Consequently, the framework is highly flexible and well-suited for modeling diverse physical systems.

\subsection{Spatio-Temporal Operator Network (STONe)}
To forecast the evolution of high-dimensional spatiotemporal systems, we adapt the S-DeepONet architecture to learn the system's evolution operator. The proposed structure is designed to approximate the mapping $\mathcal{G}_\theta$ from a history of $K_{\text{hist}}$ state observations to a sequence of $K_{\text{fut}}$ future states. The operator maps the time-dependent input history $\bm{u}_{\text{hist}}(t) = \{ \bm{y}(t - (k-1)\Delta t, \cdot) \}_{k=1}^{K_{\text{hist}}}$ to a multivariate output field $\bm{y}(\bm{r}, t_k) \in \mathbb{R}^p$ for a sequence of future times $t_k = t + k\Delta t$, where $p$ is the number of physical quantities of interest, and $k$ is a future time-step index $k \in \{1, \dots, K_{\text{fut}}\}$.

This framing is analogous to data-driven PDE discovery, where sparse identification methods are used to find parsimonious dynamical models from measurement data \cite{rudy2017data, brunton2016discovering}. In this STONe architecture, the \textbf{branch network} processes the input history $\bm{u}_{\text{hist}}$ to produce a set of time-dependent coefficients $b(\bm{u}_{\text{hist}}) \in \mathbb{R}^q$ that encode the temporal dynamics, while the \textbf{trunk network} takes as input a spatial coordinate $\bm{r}$ and a future time-step index $k$ to output a matrix of spatiotemporal basis functions $T(\bm{r}) \in \mathbb{R}^{q \times p \times K}$.

The prediction for the $p$-dimensional state vector at location $\bm{r}$ and future time $t_k$ is synthesized via a matrix-vector product. Specifically, the $j$-th component of the predicted state, $y_j$, is given by:
\begin{equation}
    y_j(\bm{r}, t + k\Delta t) = \sum_{i=1}^q b_i(\bm{u}_{\text{hist}}) T_{ijk}(\bm{r}) + \beta_j, \quad \text{for } j=1, \dots, p, \text{ and } k=1, \dots, K_{\text{fut}},
    \label{eq:sdeeponet}
\end{equation}
where $b_i$ is the $i$-th element of the branch output, $T_{ijk}$ is the element of the trunk's output matrix corresponding to the $i$-th latent dimension and $j$-th output variable at $k$-th lead time, and $\beta_j$ is a learnable bias for the $j$-th output channel.

This formulation establishes a parallel with modal decomposition techniques~\cite{schmid2010dynamic, brunton2022data}. The branch network learns to compute the optimal temporal coefficients, while the trunk network learns the optimal spatiotemporal basis functions directly from data. This end-to-end learning process bypasses the explicit eigendecompositions of traditional methods and can capture complex, nonlinear relationships. Such a generalization empowers STONe for full-rollout forecasting in non-Markovian regimes, rendering it particularly advantageous for high-dimensional spatiotemporal reconstruction, long-term dynamical forecasting, and multi-physics simulations where capturing memory-dependent evolution is essential.

\subsection{Model Architecture and Training}
\begin{figure}[!htbp] 
    \includegraphics[width=1\linewidth]{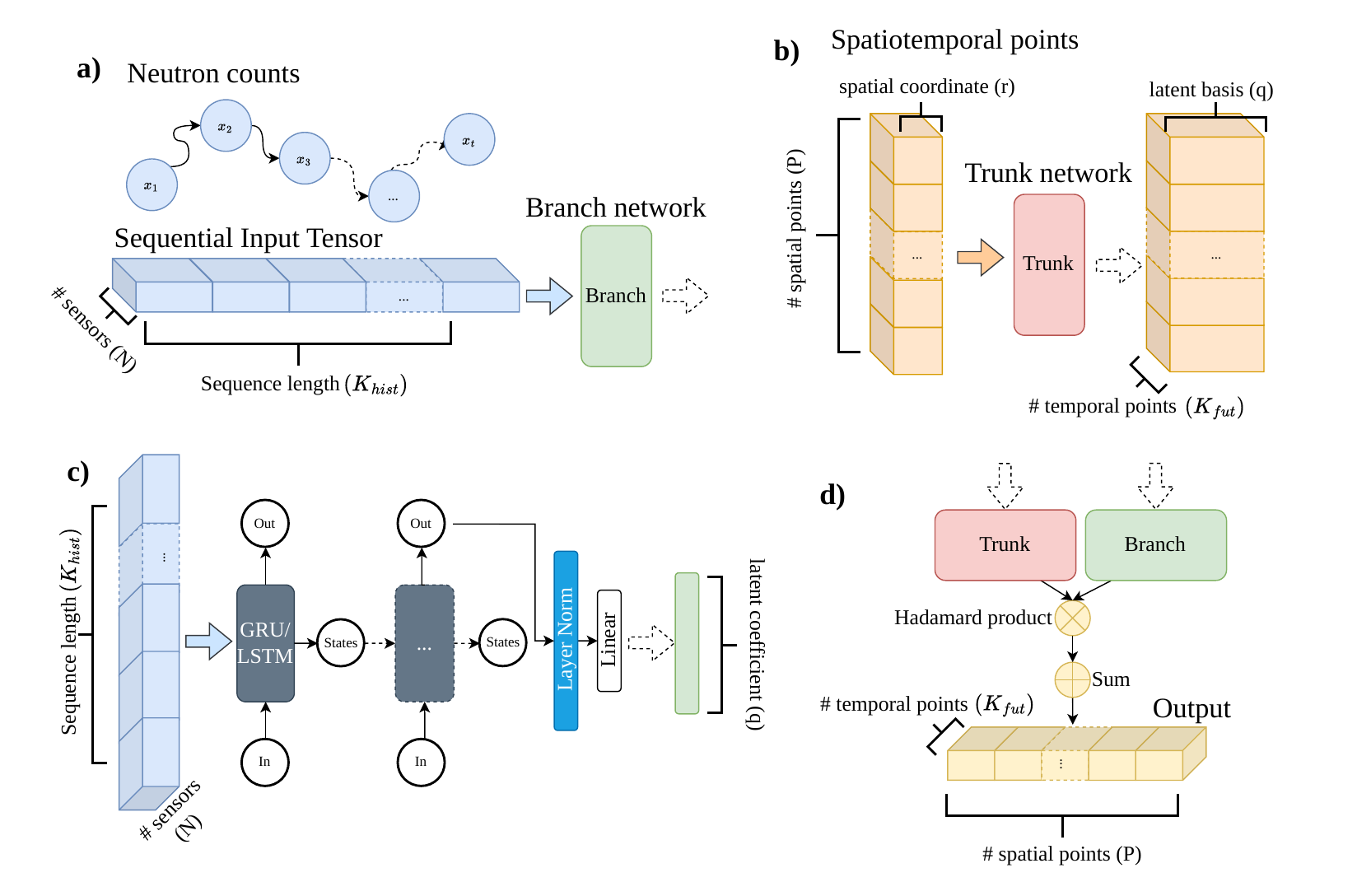}

    \caption{\textbf{STONe Architecture for Spatiotemporal Forecasting.} The modular design integrates temporal and spatiotemporal encoders where the branch network captures historical dynamics through sequence processing models, while the trunk network generates coordinate-dependent basis functions. The architecture enables principled forecasting by learning the mapping from sparse sensor histories to dense future field reconstructions. (a) Branch Network Temporal Encoder. The branch network processes sequential sensor input history $\bm{u}_{\text{hist}} \in \mathbb{R}^{K_{\text{hist}} \times N}$ through temporal encoders (GRU, LSTM, or Transformer) to extract dynamical state features and produce temporal coefficient vector $b(\bm{u}_{\text{hist}}) \in \mathbb{R}^q$. (b) Trunk Network Spatiotemporal Decoder. The trunk network maps spatiotemporal coordinates $(\bm{r})$ through fully connected layers to generate basis functions $T(\bm{r}) \in \mathbb{R}^{q \times p \times K_{fut}}$, representing the evaluation of $q$ basis functions across $p$ spacial points and $K_{fut}$ future time points. (c) Sequential Data Processing Pipeline. Historical sensor measurements from $K_{\text{hist}}$ past time steps are fed into the branch network's temporal encoder, which captures complex temporal dependencies and system dynamics to encode the current dynamical state. (d) Future State Reconstruction (Output). The STONe combines temporal coefficients from the branch network with spatiotemporal basis functions from the trunk network to reconstruct the complete future state sequence over $K_{\text{fut}}$ time steps through tensor contraction. }
    \label{fig:architecture}
\end{figure}
\vspace{-1mm}

The STONe is realized using a modular architecture where the branch and trunk networks function as temporal and spatiotemporal encoders, respectively (Figure \ref{fig:architecture}).

The \textbf{branch network} is designed to encode the temporal dependencies within the input history, $\bm{u}_{\text{hist}} \in \mathbb{R}^{K_{\text{hist}} \times N}$, where $K_{\text{hist}}$ is the number of historical time steps and $N$ is the dimension of the state vector at each step (e.g., number of sensors multiplied by the number of channels). Its objective is to map this input sequence to a latent representation of the system's current dynamical state, producing the temporal coefficient vector $b(\bm{u}_{\text{hist}}) \in \mathbb{R}^q$. To investigate different temporal feature extraction capabilities, we evaluate several architectures for this network: Gated Recurrent Units (GRUs), Long Short-Term Memory (LSTM) networks, and the Transformer architecture \cite{hochreiter1997long, chung2014empirical, vaswani2017attention}. A standard fully-connected network (FCN) that processes a flattened input vector serves as a baseline~\cite{hornik1989multilayer}. For all configurations, the core architecture consists of three hidden layers, with a latent dimension of $q=128$. For the Transformer, we employ a multi-head attention mechanism with 8 heads to effectively capture complex temporal patterns.

The \textbf{trunk network} acts as a spatiotemporal decoder, responsible for generating the basis functions upon which the dynamics evolve. It is implemented as a fully connected network with two hidden layers, each containing 128 neurons. This network maps a given spatiotemporal coordinate pair, consisting of a spatial location $\bm{r}$ and a historical time $k \in K_{hist}$, to the spatiotemporal basis matrix $T(\bm{r}) \in \mathbb{R}^{q \times p \times K_{fut}}$. This output represents the evaluation of the $q$ basis functions for each of the $p$ specific points in the future time set $K_{fut}$.

The model's trainable parameters are optimized by minimizing the mean squared error (MSE) loss function. We employ the Adam optimizer with an initial learning rate of $1 \times 10^{-3}$. To dynamically adjust the learning rate, a \texttt{ReduceLROnPlateau} scheduler was used, which reduces the learning rate by a factor of 0.5 if the validation loss does not improve by a threshold of $1 \times 10^{-4}$ for 5 consecutive epochs, down to a minimum learning rate of $1 \times 10^{-7}$. To mitigate overfitting, the training process is regularized using early stopping with a patience of 10 epochs, monitored on a held-out validation set. The model is trained for a maximum of 500 epochs on a high-performance, CUDA-enabled NVIDIA H100 GPU.

\subsection{Evaluation Metrics}
To provide a comprehensive assessment of model performance, we employ a multi-faceted evaluation protocol designed to quantify both predictive accuracy and computational efficiency. The performance of the proposed STONe architectures is systematically compared against a vanilla DeepONet baseline to rigorously evaluate the benefits of the sequential, non-autoregressive formulation.

The primary measure of forecasting accuracy is the relative L2 error ($\epsilon$), computed over the test dataset. This metric quantifies the model's ability to reconstruct the full spatiotemporal radiation field, $\bm{y}(\bm{r}, t)$, by comparing the predicted field, $\hat{\bm{y}}(\bm{r}, t)$, to the ground-truth reference from the physics-based simulations. The error is defined as:
\begin{equation}
    \epsilon = \frac{\|\bm{y} - \hat{\bm{y}}\|_{L_2}}{\|\bm{y}\|_{L_2}}
    \label{eq:l2_error}
\end{equation}
where $\|\cdot\|_{L_2}$ denotes the standard L2 norm computed across all spatial points and future time steps in the prediction horizon. To diagnose performance degradation over time, this metric is evaluated at each future lead time, $k \in \{1, \dots, K_{\text{fut}}\}$, providing insight into the model's error growth characteristics. 

Additionally, the Mean Absolute Error (MAE) and Mean Absolute Percentage Error (MAPE) are reported to assess the preservation of spatial structures crucial for identifying high-risk radiation zones and capturing the magnitude of errors in visualized model performances. The MAE is defined as:
\begin{equation}
    \text{MAE} = \frac{1}{n}\sum_{i=1}^{n}|y_i - \hat{y}_i|
    \label{eq:mae}
\end{equation}
where $n$ represents the total number of spatiotemporal points. The MAPE provides a normalized perspective on prediction accuracy:
\begin{equation}
    \text{MAPE} = \frac{1}{n}\sum_{i=1}^{n}\left|\frac{y_i - \hat{y}_i}{y_i}\right| \times 100\%
    \label{eq:mape}
\end{equation}
These complementary metrics enable comprehensive evaluation of both absolute and relative prediction errors across the spatiotemporal domain.

Beyond accuracy, the practical viability of the model for real-time applications is assessed through its efficiency. We evaluate the computational cost by measuring the total number of trainable parameters, which reflects the model's memory footprint, the computation required for its training, and the average inference time needed to generate a full-field forecast, all of which are critical for operational deployment.

This dual-focus evaluation ensures a holistic comparison, highlighting the STONe's suitability for producing high-fidelity field reconstructions in real-time, sparse-data, domain-mismatching regimes.

In addition to these quantitative metrics, we conduct a qualitative assessment to evaluate the model's ability to reproduce the physical characteristics of the radiation fields. While aggregate error metrics provide a global measure of accuracy, they can obscure localized discrepancies or a model's failure to capture critical spatial structures, such as the sharp gradients and peak intensities of high-risk radiation zones. Therefore, we perform a side-by-side visual comparison of the predicted fields against the ground-truth simulations at multiple forecast lead times. This visual analysis is crucial for verifying the structural fidelity and physical plausibility of the forecasts, ensuring that the model's predictions are not only numerically accurate but also qualitatively correct. This approach provides a more holistic understanding of model performance, complementing the quantitative results.

\subsection{Additional Validation Experiments}
\label{sec:additional_experiments}
To rigorously validate the operator-based sensing framework beyond the primary architectural comparisons, we conduct two supplementary experiments that address critical questions about sparsity limits and the necessity of operator decomposition for cross-domain virtual sensing.

\textbf{Sparsity Ablation Study.} While the primary results demonstrate reconstruction from 12 ground stations, a fundamental question remains: what is the minimum viable network density for stable cross-domain sensing? To quantify this, we perform a systematic sparsity ablation using the trained GRU-STONe model (the best-performing variant). At inference time, using the trained GRU-STONe model (our best-performing variant), we randomly mask ground stations to create progressively sparser networks of 10, 8, 6, 4, and 2 active stations. For each configuration, we evaluate on the full held-out test set over the complete 180-day horizon and compute the Relative L2 error. To account for spatial variability in which specific stations remain active, we repeat each masking configuration 5 times with different random masks and report mean error with standard deviations. This ablation quantifies how reconstruction fidelity degrades as network density decreases, establishing whether 12 stations represents a minimum threshold or a conservative operating point. 

\textbf{Classical Non-Operator Baseline.} The primary experiments compare operator-based architectures (FCN, LSTM, GRU, Transformer) within the same framework family. To establish that operator decomposition itself is essential rather than merely beneficial, we implement a classical baseline: Ridge regression with L2 regularization. This baseline represents the simplest possible cross-domain sensing model: a direct linear mapping from 12-station neutron monitor time series $\mathbf{Y} \in \mathbb{R}^{T \times 12}$ to the target dose field $\mathbf{X} \in \mathbb{R}^{T \times 65341}$ via a learned weight matrix $\mathbf{W}$: $\mathbf{X} = \mathbf{Y} \mathbf{W} + \mathbf{b}$. For the full 180-day sequence, this requires a weight matrix $\mathbf{W} \in \mathbb{R}^{2160 \times 11,761,380}$ consuming approximately 101 GB memory, exceeding our H100 GPU capacity and reflecting fundamental scaling limitations of direct sequence-to-sequence mappings. We therefore reduce the temporal window to 60 days, yielding 2.83 billion parameters with 81 GB peak memory usage. The model is trained using scikit-learn's Ridge implementation with the same train/validation/test split. Despite having 840$\times$ more parameters than GRU-STONe (3.37M), Ridge regression fails structurally: training MSE plateaus above 0.65 (vs. $<$0.004 for STONe variants), inference requires 9.177 ms (191$\times$ slower than GRU-STONe's 0.048 ms), and predictions exhibit severe spatial oversmoothing with negligible temporal structure.

\textbf{Deployable Hardware Validation.} \textbf{Deployable Hardware Validation.} The pre-trained GRU-STONe model is deployed without modification on an NVIDIA Jetson Orin Nano embedded AI platform, demonstrating physical instrument realizability beyond server-class infrastructure.
The NVIDIA Jetson Orin 
Nano is a compact embedded AI computing module based 
on the NVIDIA Ampere GPU architecture with a unified 
CPU-GPU memory system and a configurable system-level 
Thermal Design Power envelope. Its unified memory 
architecture eliminates the PCIe transfer overhead 
present in discrete GPU platforms, making it 
particularly suited for compact operator inference 
workloads. Its design targets persistent, low-power 
edge inference workloads, representative of the 
class of hardware deployable at environmental 
monitoring stations with constrained power 
infrastructure. Experiments were conducted in the 
device's ``super" power mode.

The trained GRU-STONe model (GRU branch variant, 3.37\,M parameters) was exported directly from the training environment and deployed on the Jetson Orin Nano without architectural modification, numerical retraining, weight quantization, or hardware-specific operator fusion. Inference was executed in PyTorch with CUDA acceleration on the device GPU in single-precision floating point (FP32). This direct portability without post-training adaptation constitutes a meaningful validation in itself: it confirms that the learned operator does not rely on server-class memory bandwidth, double-precision arithmetic, or high-throughput tensor core configurations that are absent on embedded platforms. System-level power consumption was recorded at 1\,Hz using the Jetson platform's onboard power monitoring interface throughout each complete 180-day rollout. Rollout latency was measured as mean wall-clock time per complete forward pass over repeated evaluations to account for CUDA kernel scheduling variability.

\section{Results}

The performance of the STONe was evaluated using four distinct temporal encoders for the branch network: FCN, LSTM, GRU, and Transformer. This section presents a comparative analysis of their forecasting accuracy and computational efficiency.

\subsection{Qualitative Results}
\begin{figure}[!htbp]
    \includegraphics[width=1\linewidth]{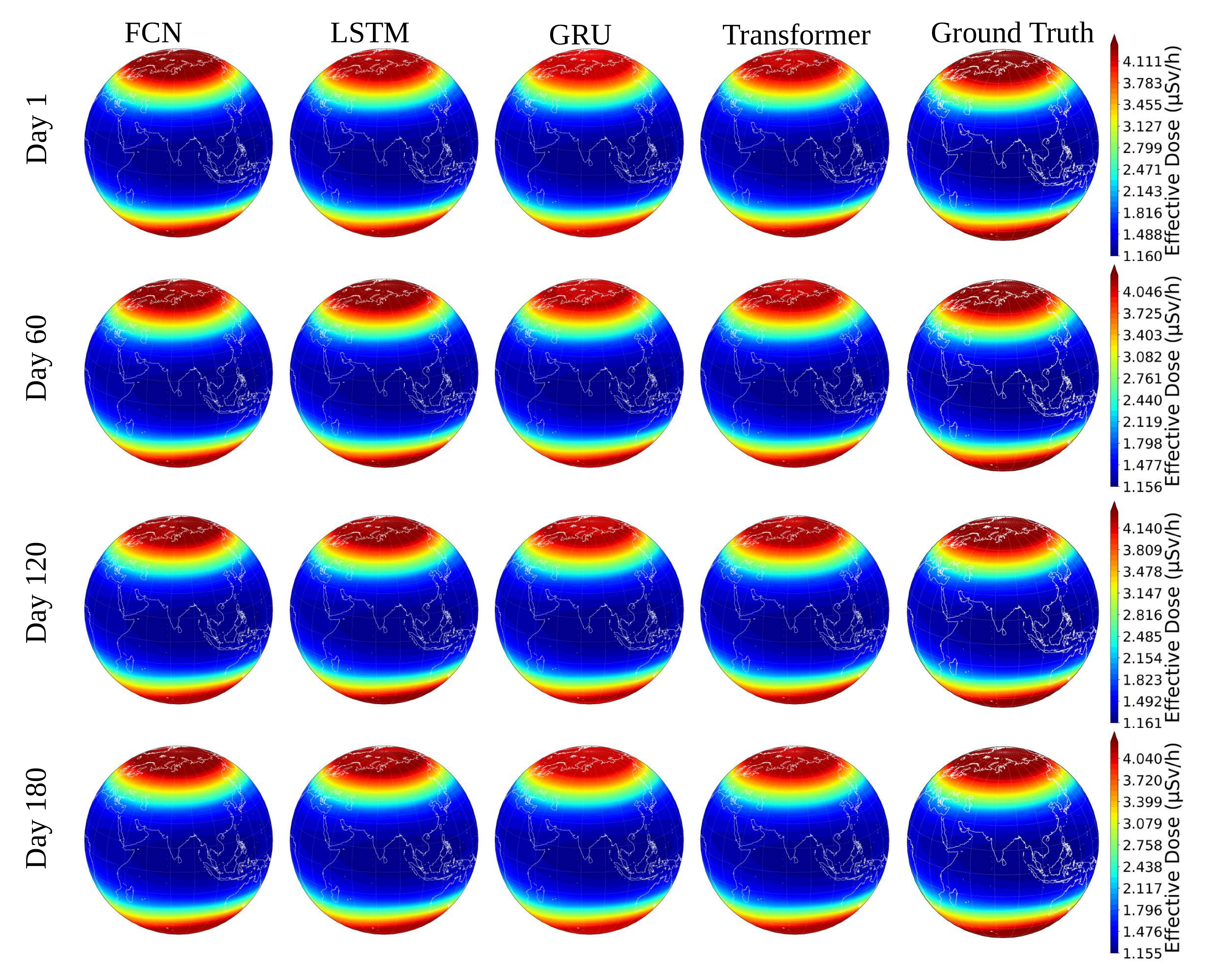}
    \caption{\textbf{Pointwise sensing error under extreme 
    constraints: MAPE fields for operator-based reconstruction 
    at aviation altitude.} Each row corresponds to a forecast 
    lead time: Day 1 (row 1), Day 60 (row 2), Day 120 (row 3), 
    and Day 180 (row 4). The maps quantify where each STONe 
    variant succeeds or fails as a \textbf{digital sensing 
    system} when reconstructing inaccessible dose fields from 
    sparse ground measurements, with elevated polar error 
    identifying the regions of greatest operational and 
    safety-critical risk.}
    \label{fig:full_comparison_grid}
\end{figure}
\vspace{-1mm}

Figure~\ref{fig:full_comparison_grid} provides a qualitative assessment of the models' spatiotemporal reconstruction capabilities at different forecast horizons (Days 1, 60, 120, and 180). The visualizations correspond to the same representative sample from the test set, allowing for direct comparison of how each model captures the evolution of the cosmic radiation field.

For effective dose forecasting, all models demonstrate competent performance, producing forecasts that are visually almost indistinguishable from the ground truth. As shown in Figure~\ref{fig:full_comparison_grid}, this demonstrates the strength of full-field reconstruction using STONe structures, capturing the location, intensity, and fine-grained spatial features of the primary radiation zones. The STONe variants incorporating sequential encoders (LSTM, GRU, and Transformer) and FCN encoder yield high-fidelity predictions, though with marginally different quantitative errors (Table~\ref{tab:performance_comparison}) and subtle visual discrepancies compared to the ground truth reconstruction.

\begin{figure}[!htbp]
    \includegraphics[width=1\textwidth, height=0.8\textheight, keepaspectratio]{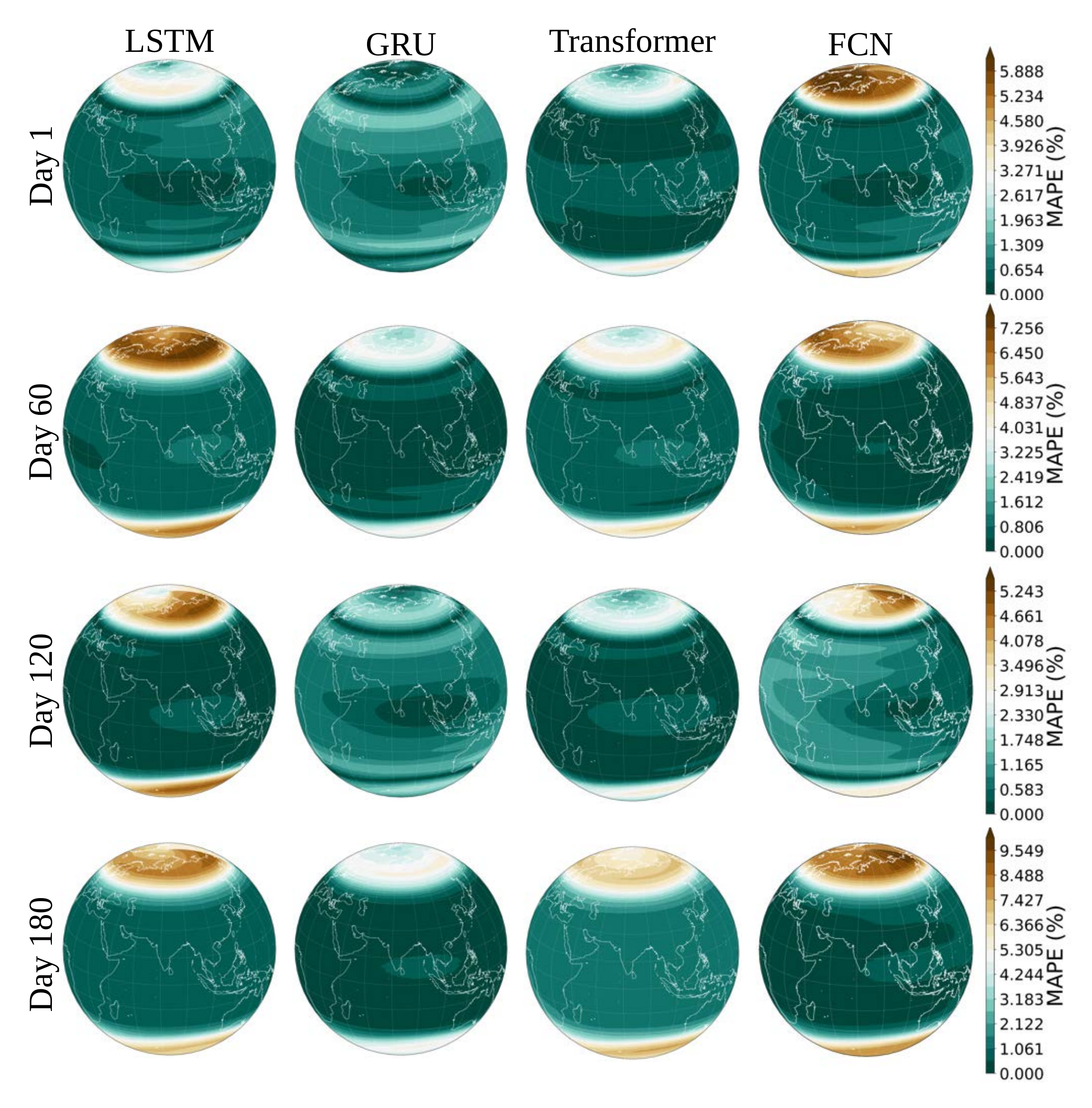}
    \caption{\textbf{Pointwise sensing error under extreme 
    constraints: MAPE fields for operator-based reconstruction 
    at aviation altitude.} Each row corresponds to a forecast 
    lead time: Day 1 (row 1), Day 60 (row 2), Day 120 (row 3), 
    and Day 180 (row 4). The maps quantify where each STONe 
    variant succeeds or fails as a \textbf{digital sensing 
    system} when reconstructing inaccessible dose fields from 
    sparse ground measurements, with elevated polar error 
    identifying the regions of greatest operational and 
    safety-critical risk.}
    \label{fig:full_comparison_error_grid}
\end{figure}
\vspace{-1mm}

Figure~\ref{fig:full_comparison_error_grid} provides insights to evaluate the superiority of variants more rigorously through visual analysis. Each row represents the forecasting results for lead times of 1, 60, 120, and 180 days, compared against the ground truth, with the MAPE calculated at each spatiotemporal point.

For near-term forecasting at Day 1, all S-DeepONet variants demonstrate superior performance compared to the vanilla DeepONet.
As the forecast horizon extends to Day 60, differences in performance become more apparent. The FCN and LSTM models begin to lose some finer spatial details, resulting in slightly blurred reconstructions in extreme regions near the poles.
As the forecast horizon extends to Days 120 and 180, the GRU and Transformer models maintain higher fidelity, preserving sharp gradients and complex shapes of the radiation patterns more effectively.

At the longest lead time (Day 180), the differences are most pronounced. The FCN's predictions, while still capturing the general location of low-intensity regions near the equator, suffer from significant oversmoothing. The GRU and Transformer models demonstrate superior performance, retaining a high degree of structural similarity to the ground truth. This visual evidence corroborates the quantitative findings, confirming that while all models are effective for short-term prediction, architectures designed to explicitly model temporal sequences (GRU and Transformer) are more robust for long-range spatiotemporal forecasting.

\begin{figure}[!htbp]
    \centering
    \includegraphics[width=1\textwidth, height=1\textheight, keepaspectratio]{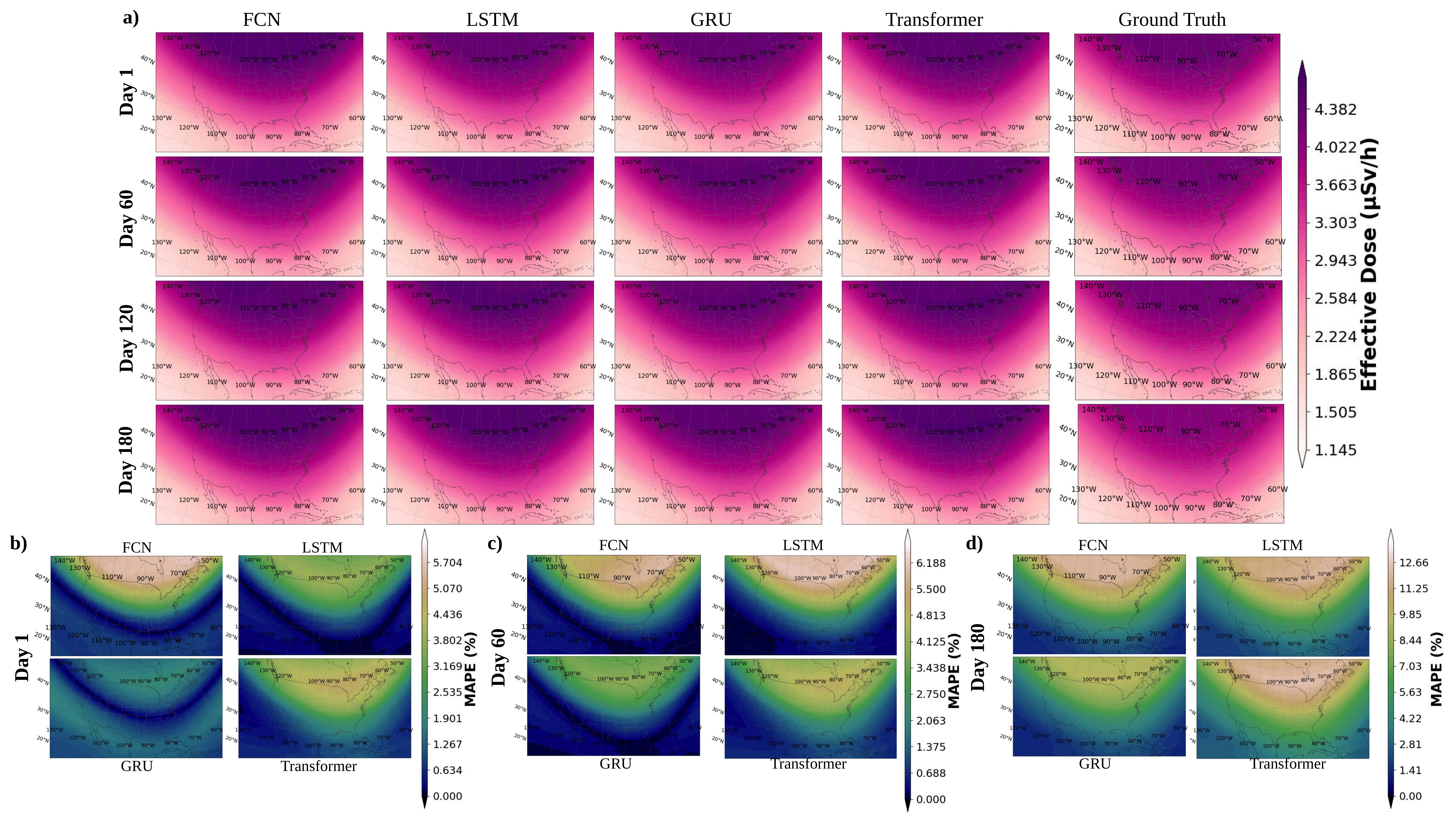}
    \caption{\textbf{Operational case study (North America): 
    stability and risk-relevant sensing performance over 180 days.}
    \textbf{(a)} Qualitative comparison of predicted effective dose 
    fields from FCN, LSTM, GRU, and Transformer models against 
    ground truth at lead times of 1, 60, 120, and 180 days (shared 
    color scale). \textbf{(b--d)} MAPE fields at Day 1, Day 60, and 
    Day 180 (independent color scales per lead time to resolve 
    relative differences). This regional view targets 
    \textbf{risk-relevant high-latitude structure} and constitutes 
    a safety-critical test of whether operator-based sensing 
    remains instrumentally reliable where occupational exposure 
    is greatest.}
    \label{fig:full_comparison_na}
\end{figure}
\vspace{-1mm}

To assess the operational utility of the proposed framework, we conduct a detailed case study focusing on the North American region, which currently experiences the highest volume of global air traffic. Figure~\ref{fig:full_comparison_na} presents a comprehensive evaluation of the four STONe branch architectures on a representative test sequence independent from the global analyses presented in previous sections.

The qualitative results (Fig.~\ref{fig:full_comparison_na}a) demonstrate that all four architectures (FCN, LSTM, GRU, and Transformer) successfully reconstruct the primary spatial features of the effective dose field across the 180-day forecast horizon. The characteristic latitudinal gradient of cosmic radiation is well-captured, and the predicted fields remain physically plausible and visually coherent even at extended lead times, validating the overall STONe framework for regional forecasting.

A quantitative analysis of the prediction error, however, reveals significant performance differences between the architectures. Figures~\ref{fig:full_comparison_na}b–d show the Mean Absolute Percentage Error (MAPE) at forecast days 1, 60, and 180. While all models perform well initially with low error across the domain (Fig.~\ref{fig:full_comparison_na}b), their long-term stability diverges considerably. By Day 180 (Fig.~\ref{fig:full_comparison_na}d), the benefits of sequential encoders become apparent. The GRU variant maintained significantly lower error, representing superior stability for long-range forecasting.

Notably, in the high-latitude northern regions, where the radiation dose is most intense and poses the greatest health risk, the GRU variant demonstrates exceptional robustness. At the 180-day lead time, it is the only architecture to maintain a MAPE below 10\% in these critical areas. In contrast, the FCN, LSTM, and Transformer models exhibit significantly higher errors, exceeding 10\% MAPE, indicating a degradation in their predictive fidelity. This high-resolution regional case study support the findings from our global analysis: while all branch structures capture the global pattern of the radiation dose, the GRU-based STONe presents the strongest and most stable performance, making it the most promising candidate for reliable, long-horizon operational forecasting.
 
\subsection{Quantitative Virtual Sensing Fidelity}

\begin{figure}[!htbp]
    \centering
    \includegraphics[width=1\linewidth]{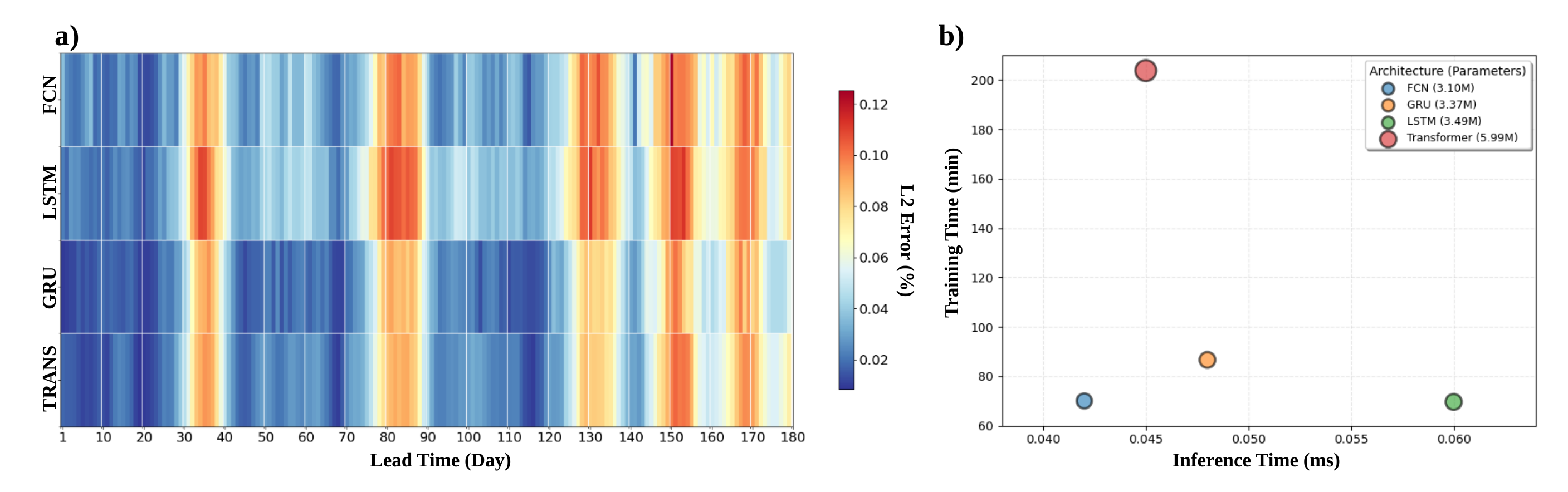}
    \caption{\textbf{Virtual sensing fidelity and system-level 
    efficiency across STONe architectures.} \textbf{a)} Relative 
    L2 error heatmap across a 180-day sensing horizon. Each row 
    corresponds to a branch architecture; colors indicate 
    instrumental error magnitude from dark blue (lowest, near 0) 
    to dark red (highest, up to 0.12). The GRU branch maintains 
    the lowest error over the full horizon, establishing the most 
    stable operator-based sensing under persistent deployment 
    conditions. \textbf{b)} Sensing system efficiency landscape 
    characterizing the trade-offs between training cost (y-axis), 
    inference latency (x-axis), and model size (bubble area, 
    parameters in millions). All variants achieve sub-millisecond 
    inference, satisfying the real-time constraint that defines 
    operational digital instrumentation; GRU provides the 
    strongest fidelity--efficiency operating point for converting 
    twelve sparse neutron monitors into a global aviation-altitude 
    virtual sensing layer.}
    \label{fig:heatmap}
\end{figure}
\vspace{-1mm}

Virtual sensing must reconstruct inaccessible fields with safety-critical fidelity
at operational latencies. We evaluate STONe across both dimensions, quantifying
(i) sensing fidelity via relative L2 error, RMSE, and MAE on held-out test data
(Supplementary Table S1), and (ii) inference latency and training requirements.

Figure~\ref{fig:heatmap}a reveals that the GRU branch sustains 
the lowest reconstruction error across the complete 180-day 
sensing horizon, establishing the most instrumentally stable 
recovery of the latent dose field under persistent deployment.  GRU shows markedly 
lower drift than FCN and LSTM, retaining long-range temporal 
structure across the 180-day horizons required for operational 
monitoring.

Figure~\ref{fig:heatmap}b characterizes inference latency and training 
requirements across all STONe variants. All architectures operate in 
the sub-millisecond regime while generating complete global field 
trajectories over 180-day horizons, indicating that the primary 
constraint for operator-based sensing is reconstruction fidelity 
under domain mismatch rather than forward-pass latency. Sub-millisecond 
inference enables closed-loop operational workflows, including 
radiation-aware routing and real-time exposure management, that remain 
infeasible with physics-based simulation regardless of computational 
resources.

\begin{table}[!htbp]
    \centering
    \caption{\textbf{Performance comparison of forecasting models across different lead times.} The table provides a comprehensive comparison of the Relative L2 Error (\%), Root Mean Squared Error (RMSE), and Mean Absolute Error (MAE) for each of the four STONe branch architectures across the 180-day forecasting lead time at 30-day intervals. The best performance for each metric at a given forecast day is highlighted in bold. }
    \label{tab:performance_comparison}
    \begin{tabular*}{\textwidth}{@{\extracolsep{\fill}}lcccc@{}}
        \toprule
        \begin{tabular}[c]{@{}l@{}}Forecast Lead \\ Time (Day)\end{tabular} & Model & Relative $L_2$ Error (\%)& RMSE & MAE \\ \midrule
        \multirow{4}{*}{1} & FCN & 0.0370 & 0.1214 & 0.0876 \\
         & LSTM & 0.0223 & 0.0732 & 0.0529 \\
         & GRU & \textbf{0.0096} & \textbf{0.0316} & \textbf{0.0261} \\
         & Transformer & 0.0178 & 0.0583 & 0.0412 \\ \midrule
        \multirow{4}{*}{30} & FCN & 0.0463 & 0.1481 & 0.1069 \\
         & LSTM & 0.0491 & 0.1570 & 0.1156 \\
         & GRU & \textbf{0.0295} & \textbf{0.0939} & \textbf{0.0695} \\
         & Transformer & 0.0395 & 0.1262 & 0.0962 \\ \midrule
        \multirow{4}{*}{60} & FCN & 0.0354 & 0.1155 & 0.0825 \\
         & LSTM & 0.0445 & 0.1450 & 0.1039 \\
         & GRU & \textbf{0.0234} & \textbf{0.0763} & \textbf{0.0552} \\
         & Transformer & 0.0241 & 0.0786 & 0.0570 \\ \midrule
        \multirow{4}{*}{90} & FCN & 0.0573 & 0.1800 & 0.1362 \\
         & LSTM & 0.0673 & 0.2119 & 0.1619 \\
         & GRU & 0.0546 & 0.1714 & 0.1306 \\
         & Transformer & \textbf{0.0513} & \textbf{0.1613} & \textbf{0.1252} \\ \midrule
        \multirow{4}{*}{120} & FCN & 0.0351 & 0.1146 & 0.0832 \\
         & LSTM & 0.0398 & 0.1300 & 0.0941 \\
         & GRU & \textbf{0.0197} & \textbf{0.0643} & \textbf{0.0467} \\
         & Transformer & 0.0259 & 0.0845 & 0.0627 \\ \midrule
        \multirow{4}{*}{150} & FCN & 0.0968 & 0.3005 & 0.2239 \\
         & LSTM & 0.0918 & 0.2849 & 0.2214 \\
         & GRU & 0.0916 & 0.2842 & 0.2209 \\
         & Transformer & \textbf{0.0885} & \textbf{0.2747} & \textbf{0.2181} \\ \midrule
        \multirow{4}{*}{180} & FCN & 0.0808 & 0.2559 & 0.1873 \\
         & LSTM & 0.0799 & 0.2529 & 0.1921 \\
         & GRU & \textbf{0.0576} & \textbf{0.1820} & \textbf{0.1368} \\
         & Transformer & 0.0731 & 0.2313 & 0.1825 \\ \midrule
        \multirow{4}{*}{\textbf{Avg.}} & FCN & 0.0523 & 0.1656 & 0.1235 \\
         & LSTM & 0.0566 & 0.1794 & 0.1348 \\
         & GRU & \textbf{0.0415} & \textbf{0.1311} & \textbf{0.0995} \\
         & Transformer & 0.0440 & 0.1389 & 0.1073 \\ 
        \bottomrule
    \end{tabular*}
\end{table}
\vspace{-1mm}

A clear pattern of error-prone regions and error accumulation is observable across all models. This behavior, where error metrics generally increase with forecast lead time, is characteristic of multi-step forecasting models, as small initial inaccuracies propagate and amplify throughout the prediction sequence. The GRU and Transformer architectures experience the least error accumulation, maintaining moderate performance levels toward the end of the 180-day forecasting period. However, the global trend of elevated errors in specific temporal regions persists across all models. Additionally, forecasts at later stages suffer more significantly from accumulated errors, as the fundamental limitation of decreasing information availability over extended horizons remains unresolved. Collectively, all performance metrics suggest that the GRU architecture possesses superior capacity for retaining long-term dependencies compared to other architectures in this cosmic muon operator learning scenario.

Table~\ref{tab:performance_comparison} provides a detailed quantitative breakdown of the models' performance at discrete intervals, reinforcing the findings from the heatmap. The GRU architecture consistently demonstrates superior performance, achieving the lowest average error across all metrics: Relative L2 Error (0.0415), RMSE (0.1311), and MAE (0.0995). The Transformer model is a strong competitor, securing the second-best average performance and even outperforming the GRU at specific intermediate forecast horizons (Day 90 and 150), which suggests a particular strength in capturing mid-range dependencies. In contrast, the FCN and LSTM models consistently lag behind, exhibiting higher error rates that indicate a greater difficulty in modeling the complex, long-term temporal patterns of the data. These quantitative results confirm that the GRU's robust performance across the majority of the forecast lead times establishes it as the most stable and accurate architecture for this task.

\subsection{Computational Efficiency}


\begin{table}[!htbp]
    \centering
    \sisetup{round-mode=places, round-precision=4, table-format=4.2}
    \caption{Computational performance and training characteristics of the different branch architectures. Inference time is the average time to generate a single 180-day forecast sequence.}
    \label{tab:efficiency}
    \begin{tabularx}{\textwidth}{X S[table-format=1.2] S[table-format=1.2] c S[table-format=1.3]}
        \toprule
        \textbf{Branch Architecture} & {\makecell{\textbf{Parameters}\\\textbf{(M)}}} & {\makecell{\textbf{Train Time}\\\textbf{(min/Epoch)}}} & {\makecell{\textbf{Train Epochs to}\\\textbf{Converge}}} & {\makecell{\textbf{Inference Time}\\\textbf{(ms)}}} \\
        \midrule
        FCN         & 3.102849 & 5.8368 & 12 & 0.042 \\
        GRU         & 3.374081 & 5.7769 & 15 & 0.048 \\
        LSTM        & 3.491329 & 5.8026 & 12 & 0.060 \\
        Transformer & 5.988865 & 5.8225 & 35 & 0.045 \\
        \bottomrule
    \end{tabularx}
\end{table}

The computational performance and training characteristics of each model architecture are detailed in Table~\ref{tab:efficiency}. The FCN is the most compact model, while the Transformer has the largest parameter count due to its multi-head attention mechanisms. Despite these differences in size, all architectures exhibit sub-millisecond inference times for a full 180-day forecast. The Transformer's highly parallelizable architecture allows its inference speed to be competitive with the smaller recurrent models (GRU, LSTM), highlighting a non-linear relationship between parameter count and inference latency.

Regarding training, the time per epoch was comparable across all architectures, averaging just under 6 minutes. However, the number of epochs required for convergence, as determined by our early stopping criteria, varied significantly. The Transformer model, with its larger parameter space, required more than double the training epochs of the recurrent models to converge. All models were trained on a single NVIDIA H100 GPU, with a maximum recorded memory usage of 26,636 MiB during the training process. Performance benchmarks for inference were conducted on the same hardware, where the reported time is an average of 100 runs over the 365-day test dataset.

A single global Monte Carlo transport evaluation requires on the order of $10^{3}$–$10^{4}$ seconds of GPU time depending on atmospheric sampling complexity, whereas STONe produces the complete 180-day global field in 0.048\,ms (GRU variant). This corresponds to a computational acceleration exceeding $10^{7}$ in wall-clock time. Given typical H100 power draw (300–350\,W), the energy required per inference is on the order of $10^{-5}$\,Wh, several orders of magnitude lower than physics-based transport. The computational barrier to operational radiation sensing is therefore resolved.

FCN achieves the lowest latency but at the cost of long-horizon 
fidelity, an instrumentally unacceptable trade-off for 
safety-critical monitoring. GRU provides the strongest 
system-level operating point, sustaining 0.048\,ms inference 
while preserving long-horizon stability. The Transformer retains 
competitive inference time despite the largest parameter count, 
confirming deployment latency is decoupled from model capacity 
in highly parallelizable architectures.

All models were trained on a single NVIDIA H100 GPU (peak memory 
26{,}636\,MiB, ~5.8 minutes per epoch), with convergence requirements 
rather than forward-pass cost governing total training time. Combined 
with the fidelity results in Figure~\ref{fig:heatmap}a, these findings 
demonstrate that operator-based virtual sensing achieves both 
reconstruction accuracy and computational efficiency necessary for 
operational deployment.

\subsection{Sparsity Ablation: Quantifying Minimum Viable Network Density}
\begin{figure}[h!]
    \centering
    \includegraphics[width=1\linewidth]{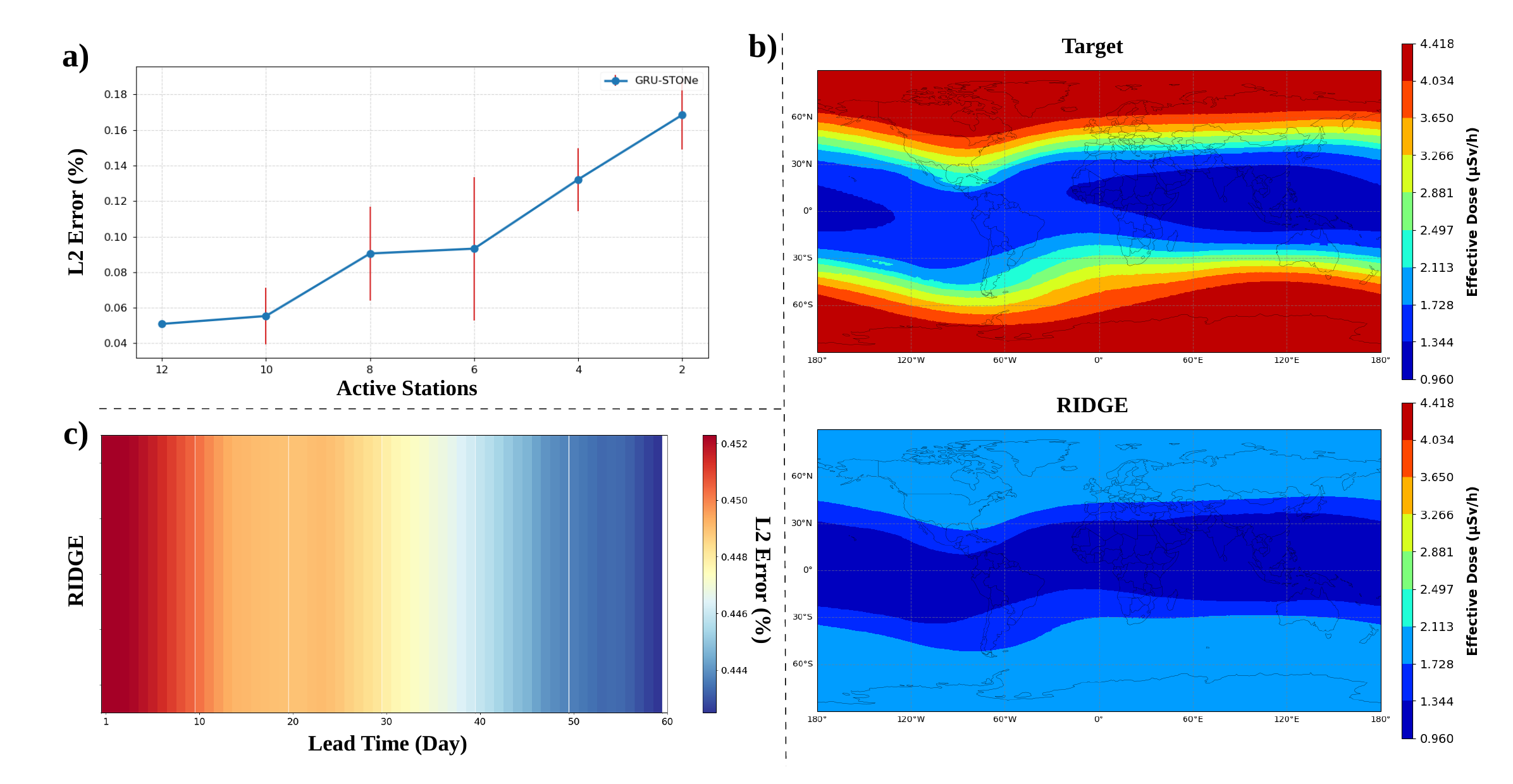}
    \caption{\textbf{Additional validation experiments for operator-based virtual sensing.} \textbf{a)} Sparsity ablation study: Relative L2 reconstruction error vs. active station count. Error bars represent standard deviation over 5 random mask configurations. Reconstruction fidelity degrades gracefully from 12 to 8 stations, then more substantially below 6 stations, with the operator retaining global sensing capability even at extreme 2-station sparsity. \textbf{b)} Ridge regression baseline failure: Top panel shows ground truth dose field at Day 1; bottom panel shows Ridge regression prediction exhibiting severe spatial oversmoothing (underfitting) despite 2.83 billion parameters. The model fails to resolve latent dose gradients or localized structures. \textbf{c)} Ridge regression error heatmap over 60-day forecast horizon. Spatially uniform high error persists across all lead times, demonstrating that classical direct mappings cannot learn physically meaningful cross-domain coupling regardless of parameter count. Operator decomposition is essential for stable virtual sensing.}
    \label{fig:additional_experiments}
\end{figure}
While the main results establish that STONe reconstructs global dose fields from 12 ground stations, a fundamental question remains: \textbf{what is the minimum viable network density for operator-based virtual sensing?} To answer this quantitatively, we systematically reduce the active station count at inference time and measure reconstruction fidelity degradation.

Figure~\ref{fig:additional_experiments}a presents the sparsity-fidelity trade-off curve. At the baseline 12-station configuration, the model achieves a mean Relative L2 error of 0.0509. When reduced to 10 stations, error increases modestly to 0.0554 ($\sigma = 0.0160$), representing only 8.8\% degradation—demonstrating graceful performance decline. Reconstruction error remains relatively stable down to 8 stations (Relative L2 $= 0.0905$, $\sigma = 0.0265$, +77.8\% degradation), indicating that the learned operator retains global sensing capability even when one-third of the ground network is removed and the configuration remains operationally viable. Performance degrades more substantially below 6 stations (Relative L2 $= 0.0933$, $\sigma = 0.0404$, +83.3\%), where increased variance reflects sensitivity to which specific stations remain active. At 4 stations, error approximately doubles (Relative L2 $= 0.1322$, $\sigma = 0.0177$, +159.7\%), indicating severe underconstraint. Even at the extreme 2-station limit, the operator produces spatially coherent global fields with Relative L2 $= 0.1685$ ($\sigma = 0.0195$, +231.0\% degradation)—a regime in which classical interpolation methods would fail completely.

The graceful degradation curve establishes that 12 stations represents a conservative operating point rather than a minimum threshold. The operator maintains instrumental utility down to 6--8 stations, suggesting robustness to station outages or strategic network reconfiguration. This quantifies the information compression capacity of learned operator-based sensing: a trained cross-domain operator extracts sufficient global structure from severely sparse indirect measurements to perform virtual instrumentation across physically inaccessible manifolds.

\subsection{Classical Baseline Comparison: Ridge Regression as Non-Operator Reference}

The main results compare operator-based architectures (FCN, LSTM, GRU, Transformer) within the same framework family. To establish that operator learning itself is essential for cross-domain virtual sensing, we implement a classical non-operator baseline: Ridge regression mapping input sensor sequences directly to output dose fields through a single linear transformation.

Unlike operator-based sensing, which evaluates the dose field at arbitrary query coordinates through the trunk network, Ridge regression hard-codes a fixed 12$\times$60 $\to$ 65,341$\times$60 mapping. The model cannot generalize to new spatial locations, different temporal horizons, or alternative grid resolutions. This architectural rigidity fundamentally limits its utility as a virtual sensing instrument, which must provide measurements at operationally relevant coordinates that may not coincide with training data.

Despite having 2.83 billion learnable parameters, 840$\times$ more than GRU-STONe (3.37M), Ridge regression fails to learn meaningful cross-domain mappings. Training MSE plateaus above 0.65, compared to sub-0.004 MSE achieved by all STONe variants within a few epochs, representing a 163$\times$ worse convergence. Figure~\ref{fig:additional_experiments}b shows the predicted dose field at Day 1: the Ridge model exhibits severe spatial oversmoothing (underfitting), failing to resolve latent dose gradients or localized high-dose structures. Figure~\ref{fig:additional_experiments}c presents error heatmaps across 60-day lead times: spatially uniform high error persists across all horizons, with negligible temporal structure in the learned mapping. Mean inference latency is 9.177 ms, approximately 191$\times$ slower than GRU-STONe (0.048 ms), while producing physically meaningless reconstructions. Furthermore, Ridge regression provides inference only at fixed grid locations, whereas GRU-STONe supports arbitrary query coordinates, enabling flexible spatial resolution for operational deployment.

\subsection{Physical Instrument Realization on 
Embedded Hardware}

The operator-theoretic virtual sensing framework claims not merely to produce accurate dose field reconstructions in a laboratory setting, but to constitute deployable digital instrumentation: a physical sensing system operable under the resource constraints of field hardware co-located with the sparse ground network it depends upon. Validating this claim requires demonstrating that the trained model executes within the power, memory, and latency envelopes of hardware suitable for permanent installation at existing neutron monitor stations; many of which operate at remote, high-latitude, or high-altitude sites with limited electrical infrastructure. We therefore deploy the pre-trained GRU-STONe model on an NVIDIA Jetson Orin Nano embedded AI platform and characterize inference performance under complete operational rollout conditions. 

Deployment performance metrics 
are summarized in Supplementary Table~\ref{tab:jetson_results}. The model 
achieves a mean rollout latency of 43.5\,ms per 
complete 180-day global field reconstruction at 
$1^\circ$ resolution, a peak GPU memory footprint 
of 143.3\,MB, and an average system power draw of 
7.3\,W (range: 6.2--9.9\,W), approximately 
44$\times$ lower than the H100 reference platform 
(${\sim}$320\,W). No thermal throttling was observed 
across repeated rollout evaluations, confirming 
sustained inference capability consistent with 
continuous monitoring deployment. The operator 
imposes no practical memory constraint on the 
embedded platform and leaves ample headroom for 
concurrent system processes. Both platforms satisfy 
the real-time inference requirement for operational 
digital instrumentation; the distinction is one of 
deployment context rather than operational 
suitability. For a monitoring system providing 
daily updates, 43.5\,ms per complete global 
reconstruction is operationally instantaneous 
regardless of platform.

Although absolute latency increases by approximately three orders of magnitude relative to the H100 platform (43.5 ms vs. 0.048 ms on H100), this gap does not constitute an operational constraint. For a continuous dose field monitoring system requiring minutely or hourly updates, 43.5 ms per complete 180-day global reconstruction remains effectively instantaneous; a rapid temporal cadence that traditional Monte Carlo methods simply cannot achieve. Both platforms satisfy the real-time inference requirement that defines operational digital instrumentation; the distinction between them is one of deployment context, server infrastructure versus field-deployable embedded hardware, rather than operational suitability.

Coupling the pre-trained GRU-STONe model with a Jetson Orin Nano establishes a self-contained virtual sensing instrument. Unlike conventional sensors reliant on hardware transducers, this device utilizes an embedded neural operator as its core sensing element to directly output calibrated spatiotemporal dose fields. Deployed without retraining, quantization, or hardware-specific modifications, the fixed FP32 weights serve as an intrinsic, invariant calibration—analogous to a physical sensor's factory calibration. Ultimately, this system transcends standard software deployment; it introduces a novel class of field-deployable edge instruments driven by an operator-theoretic measurement paradigm.

\begin{table}[h!] \centering \caption{\textbf{STONe (GRU variant) inference performance on NVIDIA Jetson Orin Nano embedded hardware vs.\ NVIDIA H100 reference platform.} All Jetson metrics are measured under complete 180-day global rollout at $1^\circ$ spatial resolution. Power values represent total system-level draw recorded via onboard power monitors. Latency is mean wall-clock time per full rollout inference over repeated evaluations.} \label{tab:jetson_results} \begin{tabular}{lcc}
\toprule \textbf{Metric} & \textbf{Jetson Orin Nano} & \textbf{H100} \\
\midrule
Mean rollout latency       & 43.5\,ms          & 0.048\,ms     \\
Peak GPU memory            & 143.3\,MB          & 26{,}636\,MB   \\
Avg.\ system power         & 7.3\,W             & ${\sim}320$\,W \\
Parameters count          & 3.37\,M            & 3.37\,M       \\
\bottomrule
\end{tabular}
\end{table} 



    

\section{Discussion}
Classical sensing frameworks assume that the quantity of interest 
is colocated with, or continuously connected to, the measurement 
device. This assumption is structurally violated in the GCR 
monitoring problem, where the surface measurement manifold and 
the aviation-altitude target field occupy physically disjoint 
domains with no intermediate regime in which conventional sensing 
operates. STONe addresses this by learning a non-autoregressive 
cross-domain operator that maps sparse, indirect surface 
measurements to a physically inaccessible field at 
sub-millisecond latency, and by instantiating that operator 
in a physically realizable device---an NVIDIA Jetson Orin Nano 
operating at 7.3\,W---that constitutes a complete sensing 
instrument deployable within the power and memory envelope 
of the ground network it depends upon. Together, these 
results demonstrate that operator learning enables virtual 
sensing across disjoint manifolds under the simultaneous 
frontier conditions defined above, and that the sensing 
principle is not merely computational but physically embodied.

STONe operates under four simultaneous conditions that define the sensing frontier:
severe sparsity (12 stations for global reconstruction), extreme inaccessibility 
(10,000 m altitude), long-horizon stability (180 days for regulatory compliance), 
and operational scale (real-time inference on both  server-class and field-deployable embedded hardware). Each condition individually 
eliminates conventional approaches; their combination requires operator-theoretic 
virtual sensing.

Leading deep learning forecasting 
models (NeuralGCM~\cite{kochkov2024neural}, 
Pangu-Weather~\cite{bi2023accurate}, 
GraphCast~\cite{lam2023learning}, 
GenCast~\cite{price2025probabilistic}, 
FourCastNet~\cite{pathak2022fourcastnet}) assume input and output fields share spatial domains. However, the input manifold of STONe and the target manifold are non-coincident in geometry, physics, and 
measurement principle. Applying field-to-field forecasting 
models to such problem is conceptually 
category-mismatched (Table~\ref{tab:model_comparison}). Multi-stage pipelines (encoding sensors then decoding fields) introduce compounded uncertainties, information loss, and degraded temporal memory from decoupled training objectives—failure modes amplified by severe sparsity, domain mismatch, and long-horizon persistence. STONe resolves these by learning the end-to-end sensing operator directly: the branch network constructs holistic temporal representations that the coordinate-conditioned trunk translates into full spatiotemporal fields without iterative feedback. This unified operator formulation is essential for stable cross-manifold sensing rather than an architectural preference.

Computational performance represents a fundamental operational threshold rather 
than incremental improvement. Monte Carlo transport requires hours to days per 
evaluation, precluding real-time operational use. Sub-millisecond inference 
enables continuous monitoring workflows including radiation-aware routing and 
real-time exposure management previously unattainable at aviation altitude.


Physical realizability distinguishes a sensing principle from a computational demonstration. Deployment of the pre-trained GRU-STONe operator on an NVIDIA Jetson Orin Nano embedded AI platform without architectural modification, weight quantization, or hardware-specific retraining. Thus the learned operator constitutes a physically realizable sensing device, not merely a software surrogate. The Jetson running GRU-STONe is a complete, self-contained instrument in the structural sense: the hardware body (Jetson Orin Nano) and the fixed sensing element (pre-trained operator weights) are inseparable in deployment, precisely as a conventional sensor module couples its transducer to a fixed internal calibration algorithm. At 7.3\,W average system power, 143.3\,MB peak GPU memory, and 43.5\,ms per complete 180-day global field rollout, the instrument operates within the envelope of photovoltaic-powered remote monitoring infrastructure. The direct portability of the pre-trained weights across computational environments mirrors the calibration portability of physical sensors across measurement sites: the operator's sensing function is fixed at training and transfers without adaptation, embodying the same architectural principle that defines conventional instrumentation. This framing is deliberate: what is demonstrated is not a model deployed on borrowed hardware, but a new category of sensing device whose measurement principle is operator-theoretic and whose deployment constraint is the power budget of field infrastructure, not the physical accessibility of the target domain.

Results indicate a physical hierarchy in temporal encoding requirements: 
short-horizon dynamics are captured by feedforward mappings (FCN), while 
extended-horizon stability requires recurrent memory (GRU) to track slowly 
varying solar and geomagnetic modulation. The performance differences thus 
reflect GCR transport timescales rather than architectural benchmarking alone.

The cross-manifold operator formulation defines a structural sensing class. Any system in which (i) measurements are available on an accessible manifold $\mathcal{H}_s$, (ii) the quantity of interest resides on a physically disjoint manifold $\mathcal{H}_t$, and (iii) the coupling between the two is dynamically stable and identifiable, admits operator-theoretic virtual instrumentation. Reactor core monitoring (boundary detectors inferring internal neutronic fields), subsurface energy systems (well logs inferring permeability and saturation fields), and structural health diagnostics (surface strain inferring internal damage states) satisfy these conditions. STONe therefore exemplifies a general sensing geometry rather than a radiation-specific architecture.

\begin{table}[!htbp]
    \centering
    \small 
    \caption{Comparison of \textbf{STONe (this paper)} with state-of-the-art data-driven weather forecasting models. The table highlights fundamental differences in problem formulation, methodology, and scale. Note that a direct numerical comparison of forecast accuracy is not applicable due to the disparate nature of the problems being solved (e.g., sparse-to-dense vs. dense-to-dense).}
    \label{tab:model_comparison}
    \begin{tabularx}{\textwidth}{@{} p{1.5cm} p{2.4cm} *{4}{>{\raggedright\arraybackslash}X} @{}}
        \toprule
        \textbf{Metrics} & \textbf{STONe} & \textbf{NeuralGCM} & \textbf{GraphCast} & \textbf{Pangu-Weather} & \textbf{FourCastNet} \\
        \midrule
        \textbf{Purpose} & Sparse-to-dense cross-domain forecasting. & Hybrid physics-ML weather climate simulation. & Dense-to-dense, medium-range weather forecasting. & Dense-to-dense, medium-range weather forecasting. & Dense-to-dense, medium-range weather forecasting. \\
        \midrule
        \textbf{Model Size} & 3.1M -- 6.0M parameters. & 11.5M -- 31.1M parameters + numerical solver. & 36.7M parameters. & ~64M parameters per base model (x4). & Not specified. \\
        \midrule
        \textbf{Training Data} & NMDB (2001--2023). & ERA5 reanalysis (1979--2019). & ERA5 reanalysis (1979--2017). & ERA5 reanalysis (1979--2017). & ERA5 reanalysis (1979--2015). \\
        \midrule
        \textbf{Training Time} & 1 -- 2 hours on a single H100 GPU. & 1 day -- 3 weeks on 16--256 TPUs. & ~4 weeks on 32 Cloud TPU v4 devices. & ~16 days on 192 V100 GPUs (per model). & 16 hours on 64 A100 GPUs. \\
        \midrule
        \textbf{Grid Resolution} & \textit{Input}: 12 points. \newline \textit{Output}: $1^{\circ} \times 1^{\circ}$ dense grid. & $0.7^{\circ}$, $1.4^{\circ}$, $2.8^{\circ}$ dense grids. & $0.25^{\circ} \times 0.25^{\circ}$ dense grid. & $0.25^{\circ} \times 0.25^{\circ}$ dense grid. & $0.25^{\circ} \times 0.25^{\circ}$ dense grid. \\
        \midrule
        \textbf{Forecasting Horizon} & 180 days (180 steps at 1-day increments) & $\le$15 days (360 steps at 1-hour increments) & 10 days (40 steps at 6-hour increments) & $\le$7 days (28 steps at 6-hour increments) & $\le$8 days (32 steps at 6-hour increments) \\
        \midrule
        \textbf{Input Variables} & 180 time steps of 12 sensor variables. & 9 atmospheric variables + extra features & 2 time steps of 227 variables & 69 atmospheric variables & 20 atmospheric variables \\
        \midrule
        \textbf{Output Variables} & 180 time steps of 1 radiation dose state. & 9 atmospheric states. & 227 atmospheric states. & 69 atmospheric states. & 20 atmospheric states. \\
        \bottomrule
    \end{tabularx}
\end{table}

\vspace{1em}
\section*{Limitations and Outlook}

STONe establishes the sensing principle; it also defines the next frontier that the principle 
opens. A first limitation concerns sparsity scaling. STONe establishes the sensing principle; it also defines the 
next frontier that the principle opens. A first limitation 
concerns sparsity scaling. The sparsity ablation study characterizes 
reconstruction fidelity down to 2 active stations, 
demonstrating graceful degradation, but does not address 
the spatial placement problem: which station configurations 
maximize sensing coverage for a given deployment budget. 
Systematic optimization of station geometry to identify 
minimum viable network configurations beyond simple random 
masking remains an important next step for deployment 
planning. Systematic sparsity ablation to identify minimum viable deployment density remains an important next step for deployment planning.
Four directions are essential for 
safety-critical deployment at scale. First, 
deterministic reconstruction must be extended 
to \textbf{probabilistic sensing}: ensembling, 
conformal prediction, or generative operator 
formulations that quantify sensing uncertainty 
under sparse coverage and rare geomagnetic 
conditions. An instrument without calibrated 
uncertainty bounds cannot be certified for 
occupational safety applications, regardless 
of its mean accuracy. Second, scaling to 
longer horizons and higher spatial resolution 
motivates continuous-time operator 
parameterizations and implicit neural field 
trunks that decouple sensing fidelity from 
discretization cost. Third, robustness to 
rare extreme events, including solar 
energetic particle events that produce 
acute dose spikes, requires physics-aware 
regularization and targeted augmentation 
with high-fidelity transport simulations 
to prevent distribution-shift failure 
modes precisely where the safety stakes 
are highest. A second limitation is the absence of explicit comparison against classical spatial-statistical baselines, such as geostatistical regression or kriging-style methods. The present study focuses on establishing the operator-theoretic sensing principle and its stability under cross-manifold conditions. Formal benchmarking against non-operator sensing approaches is reserved for dedicated sensing-method comparison studies.
 Fourth, interpretability of 
the learned cross-domain operator remains 
an open scientific opportunity: resolving 
the spatial sensitivity of the branch 
representation to individual ground 
stations would provide direct physical 
insight into the geographic structure 
of GCR coupling between the surface 
and aviation altitude, a result of 
independent scientific value beyond 
the sensing application.

These limitations define the engineering roadmap for operational 
deployment. They do not qualify the central contribution, which 
is the establishment of operator-theoretic virtual sensing as a 
new instrumentation class: a mathematical instrument that performs 
sensing across physical domains it has never occupied, validated 
under the four simultaneous conditions that define the frontier 
where classical sensing ends.
\vspace{1em}
\section*{Conclusion and Future Work}

A sensor is classically defined by physical colocation: the 
instrument occupies the domain it measures. This work establishes that this definition is not a physical 
necessity. It is a technological constraint that operator 
learning resolves. STONe demonstrates that a learned non-autoregressive 
cross-domain operator constitutes a new instrumentation class, 
one in which the instrument is a mathematical object that 
performs sensing across a physical domain it has never occupied, 
replacing hardware transducers in regimes where their deployment 
is permanently infeasible.

From \textbf{twelve} 
ground-based neutron monitors, STONe reconstructs the complete 
\textbf{global} dose field at \textbf{10{,}000\,m}, sustaining 
instrumental reliability over \textbf{180-day} horizons at 
\textbf{sub-millisecond} inference latency (GRU: Relative 
L2~$= 0.0415$). These four conditions (severe sparsity, 
extreme inaccessibility, long-horizon persistence, operational 
scale) are simultaneously present, and their simultaneous 
satisfaction is the evidence that operator-theoretic virtual 
sensing constitutes a sensing principle. By learning stable 
cross-domain mappings between physically disjoint manifolds 
without iterative feedback, STONe further resolves the two 
structural barriers that made this frontier inaccessible: 
domain mismatch between the sensing and target spaces, and 
error accumulation from autoregressive inference that has 
no physical meaning when input and output manifolds do not 
coincide. Critically, the claim that operator learning 
constitutes instrumentation is not a computational 
abstraction: the pre-trained GRU-STONe operator deployed 
without modification on an Jetson Orin Nano is the instrument. The sensing element is a fixed 
pre-trained operator; the instrument body is commodity 
embedded AI hardware; the output is a calibrated 
spatiotemporal dose field. This architecture is 
structurally identical to a conventional sensor module, 
with the transduction mechanism replaced by an 
operator-theoretic one. That substitution is the 
scientific contribution.

Future work will advance operator-based instrumentation 
along three directions required for certified 
safety-critical deployment. First, deterministic 
reconstruction will be extended to 
\textbf{probabilistic sensing} through conformal 
prediction, generative operator formulations, or 
principled ensembling, providing calibrated 
uncertainty bounds that regulatory certification 
of occupational exposure monitoring requires. 
Second, continuous-time operator parameterizations 
and implicit neural field trunks will decouple 
sensing fidelity from discretization, enabling 
higher spatial resolution and longer horizons 
without proportional memory cost. Third, 
physics-aware regularization and targeted 
augmentation with high-fidelity transport 
simulations will harden the operator against 
rare solar energetic particle events, the 
extreme conditions where instrumental 
reliability is most consequential and 
training data is most scarce.

\vspace{-1mm}
\begin{tcolorbox}[colback=gray!10!white, colframe=blue!50!black, 
title=\textbf{Key Conclusions}, coltitle=white, 
fonttitle=\bfseries]
\begin{itemize}
    \item \textbf{A new instrumentation class established:} 
    STONe proves that a learned operator bridging physically 
    disjoint manifolds constitutes digital instrumentation, 
    redefining sensing beyond the classical colocation 
    requirement and replacing unavailable physical sensors 
    in permanently inaccessible regimes.

    \item \textbf{Simultaneous proof under all four frontier 
    conditions:} From \textbf{12} ground stations, STONe 
    reconstructs \textbf{global} dose fields at 
    \textbf{10{,}000\,m} with stable \textbf{180-day} 
    horizons at sub-millisecond latency: severe sparsity, 
    extreme inaccessibility, long-horizon persistence, 
    and operational scale satisfied together for the 
    first time.

    \item \textbf{Order-of-magnitude operational threshold 
    crossed:} Sub-millisecond inference converts a 
    hours-to-days simulation process into real-time 
    digital instrumentation, enabling closed-loop 
    safety workflows that physics-based approaches 
    cannot support regardless of computational 
    resources.

    \item \textbf{Physically realized sensing device:}
    The pre-trained GRU-STONe operator deployed on 
    an NVIDIA Jetson Orin Nano constitutes a complete, 
    self-contained virtual sensing instrument operating 
    at average of 7.3\,W system power within the photovoltaic 
    envelope of remote field stations; demonstrating 
    that operator-theoretic instrumentation is not 
    a computational abstraction but a physically 
    deployable sensing device of a new category.

    \item \textbf{Universal instrumentation principle:} 
    The operator formulation 
    $\mathcal{G}: \mathcal{H}_s \rightarrow \mathcal{H}_t$, 
    $\mathcal{H}_s \cap \mathcal{H}_t = \emptyset$ 
    generalizes to reactor monitoring, structural 
    diagnostics, and subsurface systems: any domain 
    where the sensing manifold and the target manifold 
    are physically disjoint.
\end{itemize}
\end{tcolorbox}
\vspace{-1mm}

\section*{Acknowledgments}
This research is being performed using funding received from the DOE Office of Nuclear Energy’s Nuclear Energy University Program (NEUP). Award DOE DE-NE0009076.

This work leveraged Delta and DeltaAI advanced computing and data resources, funded by the National Science Foundation (awards OAC <2005572 and OAC <2320345) and the State of Illinois. Delta and DeltaAI are joint initiatives of the University of Illinois Urbana‑Champaign and NCSA.

We acknowledge the NMDB database www.nmdb.eu, founded under the European Union's FP7 programme (contract no. 213007) for providing data. Additionally, we express our gratitude to the institutions and observatories that maintain and operate the individual neutron monitor stations, whose invaluable contributions made this work possible. Athens neutron monitor data were kindly provided by the Physics Department of the National and Kapodistrian University of Athens. Jungfraujoch neutron monitor data were made available by the Physikalisches Institut, University of Bern, Switzerland. Newark/Swarthmore, Fort Smith, Inuvik, Nain, and Thule neutron monitor data were obtained from the University of Delaware Department of Physics and Astronomy and the Bartol Research Institute. Kerguelen and Terre Adelie neutron monitor data were provided by Observatoire de Paris and the French Polar Institute (IPEV), France. Oulu and Dome C neutron monitor data were obtained from the Sodankylä Geophysical Observatory, University of Oulu, Finland, with support from the French-Italian Concordia Station (IPEV program n903 and PNRA Project LTCPAA PNRA14-00091). Apatity neutron monitor data were provided by the Polar Geophysical Institute of the Russian Academy of Sciences. South Pole neutron monitor data were supplied by the University of Wisconsin, River Falls.

\bibliographystyle{unsrt}  
\bibliography{references}  

\begin{thebibliography}{10}

\bibitem{sigurdson2004cosmic}
Alice~J Sigurdson and Elaine Ron.
\newblock Cosmic radiation exposure and cancer risk among flight crew.
\newblock {\em Cancer investigation}, 22(5):743--761, 2004.

\bibitem{silverman2009medical}
Danielle Silverman and Mark Gendreau.
\newblock Medical issues associated with commercial flights.
\newblock {\em The Lancet}, 373(9680):2067--2077, 2009.

\bibitem{palmer2019ecmwf}
Tim Palmer.
\newblock The ecmwf ensemble prediction system: Looking back (more than) 25 years and projecting forward 25 years.
\newblock {\em Quarterly Journal of the Royal Meteorological Society}, 145:12--24, 2019.

\bibitem{schmid2010dynamic}
Peter~J Schmid.
\newblock Dynamic mode decomposition of numerical and experimental data.
\newblock {\em Journal of fluid mechanics}, 656:5--28, 2010.

\bibitem{tu2013dynamic}
Jonathan~H Tu.
\newblock {\em Dynamic mode decomposition: Theory and applications}.
\newblock PhD thesis, Princeton University, 2013.

\bibitem{brunton2022data}
Steven~L Brunton and J~Nathan Kutz.
\newblock {\em Data-driven science and engineering: Machine learning, dynamical systems, and control}.
\newblock Cambridge University Press, 2022.

\bibitem{brunton2016discovering}
Steven~L Brunton, Joshua~L Proctor, and J~Nathan Kutz.
\newblock Discovering governing equations from data by sparse identification of nonlinear dynamical systems.
\newblock {\em Proceedings of the national academy of sciences}, 113(15):3932--3937, 2016.

\bibitem{rudy2017data}
Samuel~H Rudy, Steven~L Brunton, Joshua~L Proctor, and J~Nathan Kutz.
\newblock Data-driven discovery of partial differential equations.
\newblock {\em Science advances}, 3(4):e1602614, 2017.

\bibitem{lu2021learning}
Lu~Lu, Pengzhan Jin, Guofei Pang, Zhongqiang Zhang, and George~Em Karniadakis.
\newblock Learning nonlinear operators via deeponet based on the universal approximation theorem of operators.
\newblock {\em Nature machine intelligence}, 3(3):218--229, 2021.

\bibitem{li2020fourier}
Zongyi Li, Nikola Kovachki, Kamyar Azizzadenesheli, Burigede Liu, Kaushik Bhattacharya, Andrew Stuart, and Anima Anandkumar.
\newblock Fourier neural operator for parametric partial differential equations.
\newblock {\em arXiv preprint arXiv:2010.08895}, 2020.

\bibitem{kobayashi2024improved}
Kazuma Kobayashi, James Daniell, and Syed~Bahauddin Alam.
\newblock Improved generalization with deep neural operators for engineering systems: Path towards digital twin.
\newblock {\em Engineering Applications of Artificial Intelligence}, 131:107844, 2024.

\bibitem{hossain2025virtual}
Raisa Hossain, Farid Ahmed, Kazuma Kobayashi, Seid Koric, Diab Abueidda, and Syed~Bahauddin Alam.
\newblock Virtual sensing-enabled digital twin framework for real-time monitoring of nuclear systems leveraging deep neural operators.
\newblock {\em npj Materials Degradation}, 9(1):21, 2025.

\bibitem{kobayashi2024deep}
Kazuma Kobayashi and Syed~Bahauddin Alam.
\newblock Deep neural operator-driven real-time inference to enable digital twin solutions for nuclear energy systems.
\newblock {\em Scientific reports}, 14(1):2101, 2024.

\bibitem{diab2025temporal}
Waleed Diab and Mohammed Al-Kobaisi.
\newblock Temporal neural operator for modeling time-dependent physical phenomena.
\newblock {\em arXiv preprint arXiv:2504.20249}, 2025.

\bibitem{nayak2025ti}
Dibyajyoti Nayak and Somdatta Goswami.
\newblock Ti-deeponet: Learnable time integration for stable long-term extrapolation.
\newblock {\em arXiv preprint arXiv:2505.17341}, 2025.

\bibitem{brandstetter2022message}
Johannes Brandstetter, Daniel Worrall, and Max Welling.
\newblock Message passing neural pde solvers.
\newblock {\em arXiv preprint arXiv:2202.03376}, 2022.

\bibitem{gupta2022towards}
Jayesh~K Gupta and Johannes Brandstetter.
\newblock Towards multi-spatiotemporal-scale generalized pde modeling.
\newblock {\em arXiv preprint arXiv:2209.15616}, 2022.

\bibitem{lippe2023pde}
Phillip Lippe, Bas Veeling, Paris Perdikaris, Richard Turner, and Johannes Brandstetter.
\newblock Pde-refiner: Achieving accurate long rollouts with neural pde solvers.
\newblock {\em Advances in Neural Information Processing Systems}, 36:67398--67433, 2023.

\bibitem{pathak2022fourcastnet}
Jaideep Pathak, Shashank Subramanian, Peter Harrington, Sanjeev Raja, Ashesh Chattopadhyay, Morteza Mardani, Thorsten Kurth, David Hall, Zongyi Li, Kamyar Azizzadenesheli, et~al.
\newblock Fourcastnet: A global data-driven high-resolution weather model using adaptive fourier neural operators.
\newblock {\em arXiv preprint arXiv:2202.11214}, 2022.

\bibitem{scarselli2008graph}
Franco Scarselli, Marco Gori, Ah~Chung Tsoi, Markus Hagenbuchner, and Gabriele Monfardini.
\newblock The graph neural network model.
\newblock {\em IEEE transactions on neural networks}, 20(1):61--80, 2008.

\bibitem{lam2023learning}
Remi Lam, Alvaro Sanchez-Gonzalez, Matthew Willson, Peter Wirnsberger, Meire Fortunato, Ferran Alet, Suman Ravuri, Timo Ewalds, Zach Eaton-Rosen, Weihua Hu, et~al.
\newblock Learning skillful medium-range global weather forecasting.
\newblock {\em Science}, 382(6677):1416--1421, 2023.

\bibitem{bi2023accurate}
Kaifeng Bi, Lingxi Xie, Hengheng Zhang, Xin Chen, Xiaotao Gu, and Qi~Tian.
\newblock Accurate medium-range global weather forecasting with 3d neural networks.
\newblock {\em Nature}, 619(7970):533--538, 2023.

\bibitem{kochkov2024neural}
Dmitrii Kochkov, Janni Yuval, Ian Langmore, Peter Norgaard, Jamie Smith, Griffin Mooers, Milan Kl{\"o}wer, James Lottes, Stephan Rasp, Peter D{\"u}ben, et~al.
\newblock Neural general circulation models for weather and climate.
\newblock {\em Nature}, 632(8027):1060--1066, 2024.

\bibitem{price2025probabilistic}
Ilan Price, Alvaro Sanchez-Gonzalez, Ferran Alet, Tom~R Andersson, Andrew El-Kadi, Dominic Masters, Timo Ewalds, Jacklynn Stott, Shakir Mohamed, Peter Battaglia, et~al.
\newblock Probabilistic weather forecasting with machine learning.
\newblock {\em Nature}, 637(8044):84--90, 2025.

\bibitem{he2024sequential}
Junyan He, Shashank Kushwaha, Jaewan Park, Seid Koric, Diab Abueidda, and Iwona Jasiuk.
\newblock Sequential deep operator networks (s-deeponet) for predicting full-field solutions under time-dependent loads.
\newblock {\em Engineering Applications of Artificial Intelligence}, 127:107258, 2024.

\bibitem{kobayashi2025proxies}
Kazuma Kobayashi, Samrendra Roy, Seid Koric, Diab Abueidda, and Syed~Bahauddin Alam.
\newblock From proxies to fields: Spatiotemporal reconstruction of global radiation from sparse sensor sequences.
\newblock {\em arXiv preprint arXiv:2506.12045}, 2025.

\bibitem{mavromichalaki2011applications}
H~Mavromichalaki, Athanasios Papaioannou, Christina Plainaki, C~Sarlanis, G~Souvatzoglou, M~Gerontidou, M~Papailiou, E~Eroshenko, A~Belov, V~Yanke, et~al.
\newblock Applications and usage of the real-time neutron monitor database.
\newblock {\em Advances in Space Research}, 47(12):2210--2222, 2011.

\bibitem{nmdb}
{NMDB Database}.
\newblock Nmdb: The neutron monitor database.
\newblock Real-Time Database for high-resolution Neutron Monitor measurements. Accessed: 2024-12-03.

\bibitem{iwamoto2022benchmark}
Yosuke Iwamoto, Shintaro Hashimoto, Tatsuhiko Sato, Norihiro Matsuda, Satoshi Kunieda, Yurdunaz {\c{C}}elik, Naoya Furutachi, and Koji Niita.
\newblock Benchmark study of particle and heavy-ion transport code system using shielding integral benchmark archive and database for accelerator-shielding experiments.
\newblock {\em Journal of Nuclear Science and Technology}, 59(5):665--675, 2022.

\bibitem{sato2015analytical}
Tatsuhiko Sato.
\newblock Analytical model for estimating terrestrial cosmic ray fluxes nearly anytime and anywhere in the world: Extension of parma/expacs.
\newblock {\em PloS one}, 10(12):e0144679, 2015.

\bibitem{sato2016analytical}
Tatsuhiko Sato.
\newblock Analytical model for estimating the zenith angle dependence of terrestrial cosmic ray fluxes.
\newblock {\em PloS one}, 11(8):e0160390, 2016.

\bibitem{expacs}
{Japan Atomic Energy Agency}.
\newblock {EXPACS}: {EX}cel-based program for calculating atmospheric cosmic-ray spectrum.
\newblock \url{http://phits.jaea.go.jp/expacs/}.
\newblock Accessed: November 4, 2024.

\bibitem{chen1995universal}
Tianping Chen and Hong Chen.
\newblock Universal approximation to nonlinear operators by neural networks with arbitrary activation functions and its application to dynamical systems.
\newblock {\em IEEE transactions on neural networks}, 6(4):911--917, 1995.

\bibitem{hochreiter1997long}
Sepp Hochreiter and J{\"u}rgen Schmidhuber.
\newblock Long short-term memory.
\newblock {\em Neural computation}, 9(8):1735--1780, 1997.

\bibitem{chung2014empirical}
Junyoung Chung, Caglar Gulcehre, KyungHyun Cho, and Yoshua Bengio.
\newblock Empirical evaluation of gated recurrent neural networks on sequence modeling.
\newblock {\em arXiv preprint arXiv:1412.3555}, 2014.

\bibitem{vaswani2017attention}
Ashish Vaswani, Noam Shazeer, Niki Parmar, Jakob Uszkoreit, Llion Jones, Aidan~N Gomez, {\L}ukasz Kaiser, and Illia Polosukhin.
\newblock Attention is all you need.
\newblock {\em Advances in neural information processing systems}, 30, 2017.

\bibitem{hornik1989multilayer}
Kurt Hornik, Maxwell Stinchcombe, and Halbert White.
\newblock Multilayer feedforward networks are universal approximators.
\newblock {\em Neural networks}, 2(5):359--366, 1989.

\end{thebibliography}

\end{document}